\documentclass[sigconf]{acmart}
\usepackage{soul}
\usepackage{booktabs} 
\usepackage{graphicx}
\usepackage{subfigure}
\usepackage{amsthm}
\usepackage{amsmath}
\usepackage{multirow}
\usepackage{enumerate}
\usepackage{amsfonts}
\usepackage{color}
\usepackage{url}
\usepackage{amsfonts}
\usepackage{tabularx}
\usepackage{diagbox}
\usepackage{longtable}
\usepackage{rotating}
\usepackage{slashbox}
\usepackage[justification=centering]{caption}
\usepackage[ruled, linesnumbered]{algorithm2e}
\usepackage{bm}
\usepackage{float}
\usepackage{hyperref}
\def\model{NEAT}

\AtBeginDocument{%
  }

\setcopyright{acmcopyright}
\copyrightyear{2024}
\acmYear{2024}
\acmDOI{XXXXXXX.XXXXXXX}

\acmConference[Conference acronym 'XX]{Make sure to enter the correct
  conference title from your rights confirmation emai}{June 03--05,
  2018}{Woodstock, NY}
\acmPrice{15.00}
\acmISBN{978-1-4503-XXXX-X/18/06}

\begin{document}
\settopmatter{printacmref=false} 
\renewcommand\footnotetextcopyrightpermission[1]{} 
\pagestyle{empty} 
\title{Unsupervised Generative Feature Transformation via Graph Contrastive Pre-training and Multi-objective Fine-tuning}


\author{Wangyang Ying}
\affiliation{%
  \institution{Arizona State University}
  \city{Tempe}
  \state{Arizona}
  \country{USA}
}\email{wangyang.ying@asu.edu}

\author{Dongjie Wang}
\affiliation{%
  \institution{University of Kansas}
  \city{Lawrence}
  \state{Kansas}
  \country{USA}}
\email{wangdongjie@ku.edu}

\author{Xuanming Hu}
\affiliation{%
  \institution{Arizona State University}
  \city{Tempe}
  \state{Arizona}
  \country{USA}
}\email{solomonhxm@asu.edu}

\author{Yuanchun Zhou}
\affiliation{%
 \institution{Computer Network Information Center, Chinese Academy of Sciences}
 \city{Beijing}
 \country{China}}
\email{zyc@cnic.cn}

\author{Charu C. Aggarwal}
\affiliation{%
 \institution{IBM T. J. Watson Research Center}
 \city{Yorktown Heights}
 \state{New York}
 \country{USA}}
\email{charu@us.ibm.com}

\author{Yanjie Fu$^\dagger$}
\affiliation{%
 \institution{Arizona State University}
 \city{Tempe}
 \state{Arizona}
 \country{USA}}
\email{yanjie.fu@asu.edu}
\thanks{$\dagger$ Corresponding Author}

\renewcommand{\shortauthors}{Trovato et al.}

\begin{abstract}
Feature transformation is to derive a new feature set from original features to augment the AI power of data.
In many science domains such as material performance screening, 
while feature transformation can model material formula interactions and compositions and discover performance drivers, supervised labels are collected from expensive and lengthy experiments. 
This issue motivates an Unsupervised Feature Transformation Learning (UFTL) problem.
Prior literature, such as manual transformation, supervised feedback guided search, and PCA, either relies on domain knowledge or expensive supervised feedback, or suffers from large search space, or overlooks non-linear feature-feature interactions.  
UFTL imposes a major challenge on existing methods: how to design a new unsupervised paradigm that captures complex feature interactions and avoids large search space?
To fill this gap, we connect graph, contrastive, and generative learning to develop a measurement-pretrain-finetune paradigm for UFTL. 
For unsupervised feature set utility measurement,  we propose a feature value consistency preservation perspective and develop a mean discounted cumulative gain like unsupervised metric to evaluate feature set utility. 
For unsupervised feature set representation pretraining, we regard a feature set as a feature-feature interaction graph, and develop an unsupervised graph contrastive learning encoder to embed feature sets into vectors. 
For generative transformation finetuning, we regard a feature set as a feature cross sequence and feature transformation as sequential generation. We develop a deep generative feature transformation model that coordinates the pretrained feature set encoder and the gradient information extracted from a feature set utility evaluator to optimize a transformed feature generator. 
Finally, we conduct extensive experiments to demonstrate the effectiveness, efficiency, traceability, and explicitness of our framework. Our code and data are available at \url{https://shorturl.at/pKQU5}.  
\end{abstract}



%
\maketitle
\makeatletter
\renewcommand{\maketag@@@}[1]{\hbox{\m@th\normalsize\normalfont#1}}%
\makeatother

\vspace{-0.1cm}
\section{Introduction}


Feature transformation (FT) is to derive new features from original features to reconstruct a transformed feature space (e.g. $[f_1, f_2] \rightarrow [\frac{f_1}{f_2}, f_1-f_2, \frac{f_1+f_2}{f_1}]$). 
As an essential task of data-centric AI, FT is practical and effective in industrial deployments, because it can augment the AI power of data (e.g., structural, predictive, interaction, and expression levels) to achieve better predictive performance even with simple models. 
In many practices, FT is conducted either by human experts, or by machine-assisted search guided through downstream task feedback. 
However, in certain domains, such as material performance screening, although FT can model material formula interactions and compositions to discover material performance drivers, supervised labels are collected from lengthy and expensive experiments. 
This issue motivates a new learning problem: Unsupervised Feature Transformation Learning (UFTL). 
Solving UFTL can reduce data costs, model complex feature-feature interactions, learn from non-labeled data, and improve model generalization capability.  

Existing systems only partially solve UFTL: 
1) manual transformations are effective and explicit, but not generalizable and incomplete, as they heavily rely on domain and empirical experiences. 
2) supervised transformations include exhaustive-expansion-reduction approaches~\cite{expansion-reduction1, expansion-reduction2, expansion-reduction3, expansion-reduction4, expansion-reduction5} and iterative-feedback-improvement approaches~\cite{envolution1, envolution2, envolution3} that are enabled by reinforcement, genetic algorithms, or evolutionary algorithms. Such methods can search optimal feature sets in a vast discrete space, but they face exponentially growing combination possibilities, are computationally costly, hard to converge, and unstable. Moreover, they rely on labeled data, which limits their applicability. 
3) unsupervised transformations, such as Principal Components Analysis (PCA)~\cite{PCA,uddin2021pca}, don't rely on labeled data. But, PCA is based on a strong assumption of straight linear feature correlation, only uses two cross operations (i.e., addition/subtraction) to generate new features, and only reduces (can't increase) feature space dimensionality. 


\begin{figure}[!t]
        \centering
\includegraphics[width=1.0\linewidth]{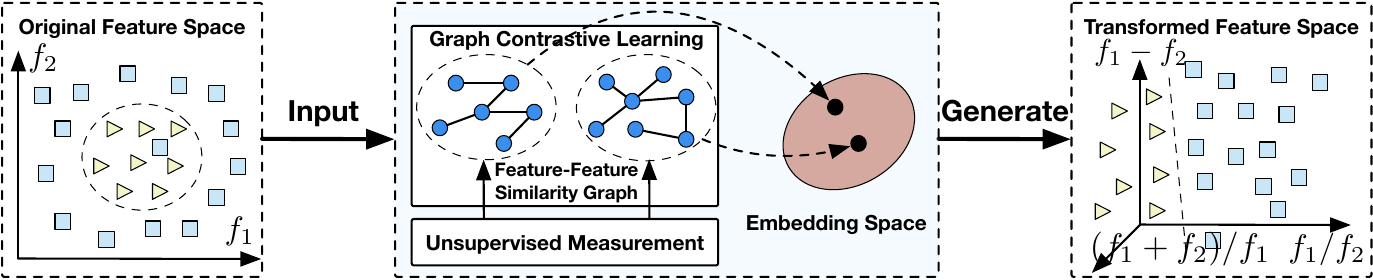}
	\caption{Unsupervised Feature Transformation Learning.}
        \vspace{-0.5cm}
	\label{motivation}
\end{figure}
The emergence of LLM (e.g., ChatGPT) shows that mechanism-unknown human language knowledge can be embedded into a large embedding space and model the world as generative AI to autoregress the next word. 
Following a similar spirit, we believe that mechanism-unknown feature space knowledge can be embedded into large sequential foundational models for generating a transformed feature set.
For example, a transformed feature set ({$\frac{f_1}{f_2}, f_1-f_2, \frac{f_1+f_2}{f_1}$}) is seen as a feature cross sequence ``$f_1/f_2, f_1-f_2, (f_1+f_2)/{f_1}$EOS''. 
This provides a great potential to transform the traditional ways that we solve UFTL. 

\textbf{Our contribution: a graph, contrastive, and generative learning perspective.} 
We connect graph, contrastive, and generative learning to develop a measurement-pretrain-finetune paradigm, as shown in Figure~\ref{motivation}.
Specifically, 
1) \textit{unsupervised feature set utility measurement:} Under the unsupervised setting, downstream task feedback and labeled data are not available for measuring feature set utility. We propose a feature value consistency preservation angle. The insight is that: if two data instances (i.e., rows) are similar, both instances tend to be from the same class, therefore the two element values of a specific feature over the two instances should be close to each other. 
We reformulate feature set utility as information gain of total feature value consistency from the lens of instance pair similarity.  
2) \textit{unsupervised graph contrastive pretraining:} Aside from feature set utility, feature set representation needs to be learned in an unsupervised fashion. We see a transformed feature set as a feature-feature similarity graph, and reformulate the feature set2vec problem into an unsupervised graph contrastive learning task. 
By seeing a feature set as a graph rather than just a feature cross sequence, the graph contrastive pretraining can better model feature-feature interactions, learn a feature set level embedding space, and accelerate later finetuning and training convergence.
3) \textit{multi-objective finetuning: } 
We regard a feature set as a feature cross sequence, and feature transformation as sequential feature cross generation. 
We find that: (i)  deep sequential model can distill feature knowledge in the continuous embedding space, in order to drive autoregressive generation of transformation sequences; 
(ii) navigating optimization and generation in the continuous embedding space via gradient ascents can convert classic discrete search formulation into differentiable continuous optimization, and avoid large discrete search space.

\textbf{Summary of technical solution.} 
Inspired by these insights, this paper presents a principled and generic framework for the UFTL problem by the measurement-pretrain-finetune paradigm. 
Specifically, given an original dataset, we can explore various feature combinations of original features to prepare different feature sets. 
In \textit{the measurement stage}, we generalize the idea of mean discounted cumulative gain to feature set utility measurement. 
We see  ``information gain'' as the value consistency of a specific feature between two close data instances,  and ``cumulative'' as the sum of the consistency gains of all data instance pairs. 
``discounted'' means that: when a feature is of high variance, its element values tend to be different. Under such a situation, if the element values of such a feature over any two close instances still stay consistent,  the cumulative gain of such a high-variance feature should be augmented. 
``mean'' means that we average the cumulative discounted gain of all features as the final utility of a feature set. 
In \textit{the pretrain stage}, we regard a feature set as a feature-feature similarity graph and develop a contrastive GNN to learn feature set embedding. In particular, to prepare contrastive pairs, we apply attribute mask and edge perturbation to feature-feature similarity graphs to generate augmented graphs. 
An augmented graph will pair with a source augmented graph as a positive pair; an augmented graph will pair with a non-sourced augmented graph as a negative pair.
We optimize the contrastive loss by enforcing that the embeddings of two augmented graphs from the same source graph should be similar, and different otherwise. 
In \textit{the fine-tune stage}, we develop a deep generative feature transformation model. This model includes an encoder, a decoder, and an evaluator. 
The set2vec encoder is pretrained by and loaded from the pretrain stage to map a feature set to an embedding vector; the vec2seq decoder is to decode an embedding vector to generate a feature cross sequence that represents a feature set; the evaluator that predicts the feature set utility given feature set embedding can provide gradient-steered optimization to find the optimal transformed feature set embedding, by jointly optimizing the losses of decoder and evaluator. 


\vspace{-0.1cm}
\section{Problem Statement}

\noindent\textbf{Reinforcement Learning (RL) based Data Collector.}
The success demonstrated by ~\cite{envolution1} highlights that feature learning knowledge can effectively be modeled using multi-agent systems.
Thus, we employ its framework to automatically collect high-quality training data using our defined unsupervised feature set utility measurement for transformation embedding space. More details about the data collector are described in Appendix~\ref{appendix-RL}.



\begin{figure}[h]
    \centering
       \vspace{-0.2cm}
    \includegraphics[width=1.0\linewidth]{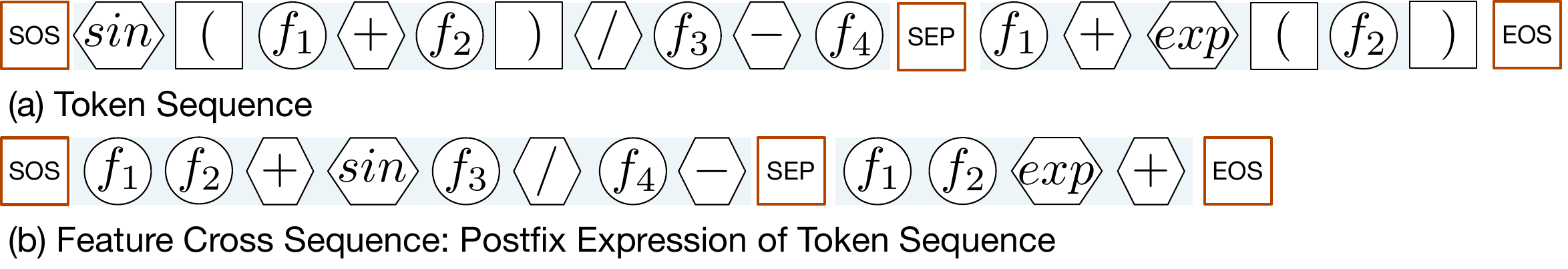}
   \vspace{-0.5cm}
    \caption{Postfix Expression.}
    \label{postfix}
    \vspace{-0.2cm}
\end{figure}

\noindent\textbf{Important Definitions.}
\ul{\emph{1) Feature Cross}}, which means we apply mathematical operations to original features to generate a new feature (e.g., $[sin(f_1+f_2)/f_3-f_4]$). 
\ul{\emph{2) Explored Feature Set}}. We combine multiple feature crosses to construct an explored feature set (e.g., $[sin(f_1+f_2)/f_3-f_4, f_1+exp(f_2)]$).
\ul{\emph{3) Feature Cross Sequence}}. Figure~\ref{postfix}(a) shows that we use a token sequence to represent the explored feature set. <SOS>, <SEP>, and <EOS> indicate the start, separate, and end tokens respectively. 
To use fewer tokens and reduce learning difficulty, this sequence can be represented by a postfix expression, as shown in Figure~\ref{postfix}(b).
In this paper, the term "feature cross sequence" refers to the representation of a transformed feature set using a postfix expression of token sequence.


\noindent\textbf{The UFTL Problem.}
Formally, given a data set $D = (X, y)$ and an operation set $\mathcal{O}$, where $X$ is an original feature set and $y$ is the corresponding label. 
Here, $y$ is only used for testing purposes and is not involved in the training process.
We first apply the RL-based data collector on $X$ to collect $N$ training data samples, denoted by $\Tilde{D} = \{\mathcal{F}_i, \mathcal{T}_i, s_i\}_{i=1}^N$, where $\mathcal{F}_i$, $\mathcal{T}_i$, and $s_i$ indicate the explored feature set, feature cross sequence, and feature set utility score of the $i$-th sample respectively.
We aim to 1) embed explored feature sets into a differentiable continuous space and 2) generate the optimal transformed feature set. 
To accomplish Goal 1, we designed an encoder-evaluator-decoder framework that compresses the feature learning knowledge of $\Tilde{D}$ into an embedding space $\mathcal{E}$, optimized through pre-training and fine-tuning processes.
To achieve Goal 2, we adopt a gradient-steered search method to find the optimal embedding point $E^*$ and reconstruct the best feature space $\mathcal{F}^*$.
It can be formulated by
\begin{equation}
    \mathcal{F}^* = \psi(E^*)[X] = \arg\max_{E \in \mathcal{E}}\mathcal{S}(\psi(E)[X]),
\end{equation}
where $\psi$ is the well-trained decoder that can regenerate the feature cross sequence from any learned embedding $E$;
$\mathcal{S}$ is the function for calculating
feature set utility; $\psi(E)[X]$ means applying the reconstructed sequence on $X$ to obtain the transformed feature set.

\section{The Measurement-Pretrain-Finetune Paradigm}

\subsection{Framework Overview}
\begin{figure*}[]
        \centering
\includegraphics[width=1.0\linewidth]{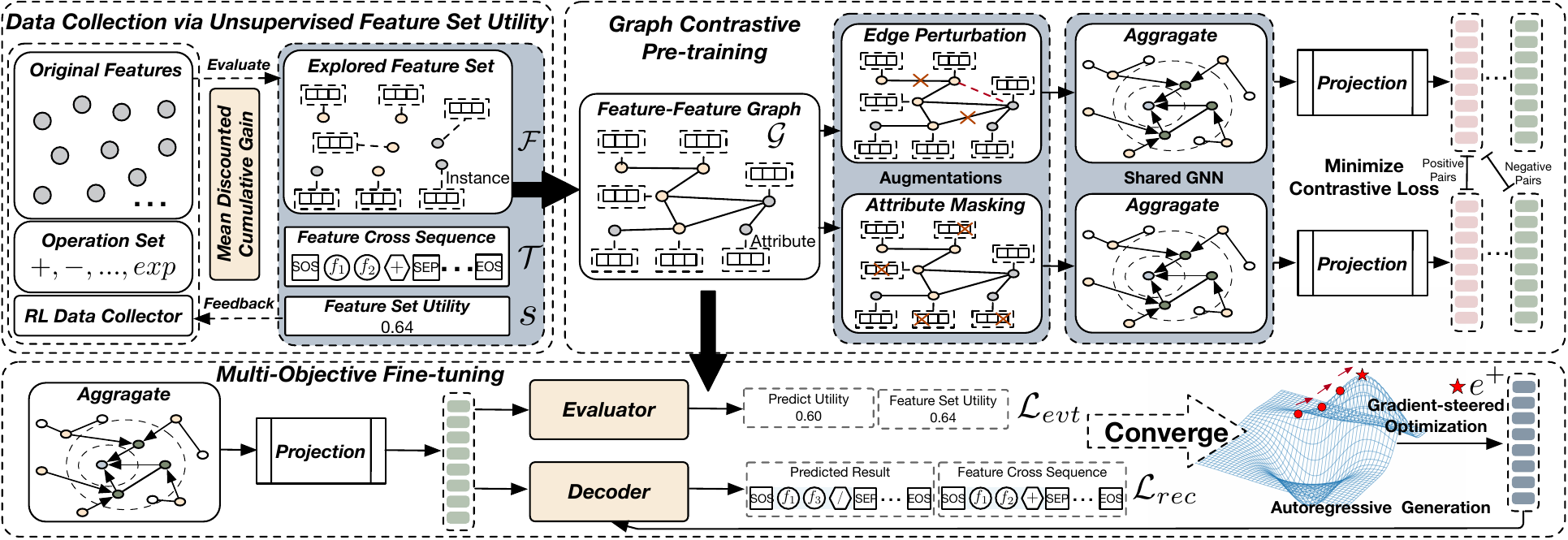}
 \vspace{-0.3cm}
	\caption{An overview of {\model}. First, we explore the original feature set to collect training data under the guidance of our proposed unsupervised feature set utility. Second, we pre-train shared GNNs to capture the knowledge of the training data via graph contrastive learning and preserve it in an embedding space. Finally, we conduct multi-objective fine-tuning to readjust the created embedding space and identify the optimal transformed feature set.}
        \vspace{-0.4cm}
	\label{framework}
\end{figure*}

Figure~\ref{framework} show our u\underline{N}supervised g\underline{E}nerative fe\underline{A}ture \underline{T}ransformation framework (\textbf{\model}) includes three steps: 
1) Data collection via unsupervised feature set utility, 
2) Graph contrastive pre-training, 
and 3) Multi-objective fine-tuning.
In Step 1, we design an unsupervised measurement to assess feature set utilities. Under the guidance of this measurement, we employ the RL-based data collector to explore the original feature space. The collector will automatically gather explored feature sets, feature cross sequences, and feature set utility for training data preparation.
In Step 2, we pre-train shared GNNs to capture the knowledge of the training data via graph contrastive learning and preserve it in an embedding space. 
In detail,  we first extract feature-feature attributed graphs from the explored feature set.
Then, sampling-based GNNs are employed to capture the characteristics of these graphs through contrastive learning.
In Step 3, we conduct multi-objective fine-tuning to update the created embedding space and identify the optimal transformed feature space.
More specifically, to update the embedding space, we load the pre-trained GNNs as an encoder and create a decoder and an evaluator for feature cross sequence reconstruction and utility score estimation respectively.
Then, we search for enhanced embeddings within the new embedding space and reconstruct feature cross sequences.
Finally, these sequences are used to transform the original feature space.
The transformed feature set yielding the highest utility score is output as the optimal result.

\subsection{Unsupervised Feature Set Utility Measurement}
\label{unsupervised_measurement}

\noindent\textbf{Why measuring unsupervised feature set level utility matters.} 
Feature set utility measurement reflects the quality of the feature space, which is critical for guiding the search toward the optimal feature set.  
Conventional strategies leverage supervised methods (e.g., decision trees, feature relevance) to measure individual-level feature importance and aggregate them to set-level utility.
However, such strategies are unsuitable for the UFTL setting where downstream predictive task labels are unavailable or expensive.
Note that we can use unsupervised measures, such as feature redundancy, to evaluate individual-level feature utility, and then perform average aggregation. But such a method only evaluates the negative (e.g., information overlap) aspect of a feature, ignores the meaningful interactions of feature-instance, instance-feature value, feature value-feature vector,  and fails to consider self-weighted or self-penalized aggregation when propagating information score from individual level to set level.

\noindent\textbf{Mean Discounted Cumulative Gain (MDCG): a feature value consistency preservation perspective.}
To overcome these limitations,
we propose a feature value consistency preservation angle that illustrates an important intuition: if two data instances (i.e., rows) are similar, then both instances tend to be from the same class, thereafter the two element values of a specific feature with respect to the two similar instances should be similar. We see the alignment with such intuition as information gain and develop an MDCG based unsupervised measure. 

\noindent\ul{\emph{Step 1: Information Gain.}}
We quantify the information gain of a feature as the score of alignment with the intuition: a feature's element values with respect to two similar instances should be similar. 
Formally, given a feature set $\mathcal{F} \in \mathbb{R}^{n \times m}$, $n$ and $m$ are the number of instances and features respectively. Let $i$ and $j$  are indexes of instances, the information gain of the $q$-th feature over the $i$-th and $j$-th instances is given by:
$
    g^{q}_{ij} = {(\mathcal{F}_{iq}-\mathcal{F}_{jq})^2}{e^{-\frac{||\mathcal{F}_{i,*}-\mathcal{F}_{j,*}||^2}{constant}}}
$. 
In the experiments, we set the value of \textit{constant} as 2.
Here, the lower the value of $g^{q}_{ij}$, the greater the information gain observed.

\noindent\ul{\emph{Step 2:  Cumulative Gain.}}
The cumulative gain of the $q$-th feature over all the instance pairs is given by $\sum_{ij}{S_{ij}*g^{q}_{ij}}$, where $S_{ij}$ is a binary indicator: 1) when the $i$(or $j$)-th instance is among the K nearest neighbors of the $j$(or $i$)-th instance, $S_{ij}=1$ indicates the $i$-th instance and the $j$-th instance are similar; 2) otherwise,  $S_{ij}=0$ indicates the two instances are not similar. K is a hyperparameter given by users. 


\noindent\ul{\emph{Step 3: Discounted Cumulative Gain.}} 
When a feature exhibits high variance and its element values tend to diverge, if the element values of such feature over any two similar instances still stay similar, the cumulative gain of such high-variance feature should be augmented. Therefore, we propose the concept of discounted cumulative gain: $\frac{\sum_{ij}{S_{ij}*g^{q}_{ij}}}{Var(\mathcal{F}_{*,q})}$ to describe feature importance. 

\noindent\ul{\emph{Step 4: Mean Discounted Cumulative Gain.}} 
Finally, we quantify the feature set utility as the average score of the discounted cumulative gains of all the features: 
$
\sum_{q=1}^m(1 - \frac{\sum_{ij}{S_{ij}*g^{q}_{ij}}}{Var(\mathcal{F}_{*,q})})/m.
$

Leveraging the utility of the feature set, we employ the RL-based collector~\cite{envolution1} to gather explored feature sets $\mathcal{F}$, feature cross sequences $\mathcal{T}$, and feature set utilities $s$ for training data preparation.

\subsection{Unsupervised Feature-Feature Interaction Graph Contrastive Pre-Training}
\noindent\textbf{Why pre-training a feature set embedding space matters.} 
After collecting the training data, learning the embedding space of the feature set becomes essential for two reasons: 
1) feature set2vec can learn a continuous embedding space, so we can convert classic discrete search into effective gradient ascent in continuous optimization. 
2) feature set2vec can develop a database of feature set embedding vectors along with corresponding unsupervised utilities that enable the learning of a feature set utility evaluator. This evaluator can model the relationship between feature set embeddings and feature set utilities and provide gradient information to search for optimal feature sets. 

\noindent\textbf{Leveraging graph contrastive learning for unsupervised feature set2vec.} 
Feature set2vec, particularly under an unsupervised setting, is challenging. We propose a graph perspective to regard a feature set as a feature-feature similarity graph. This feature-feature graph can describe the geometric topology of a feature space, and convert set2vec into graph2vec, so graph contrastive learning can be introduced to achieve unsupervised embedding pretraning. 
We pre-train a GNN-based encoder to convert explored feature sets to vectors, thereby constructing a feature transformation embedding space. During the pre-training, we use the explored feature sets collected in the previous step to construct embedding space. To simplify notations, we use $\mathcal{F}$ to represent any explored feature set.

\noindent\ul{\emph{Creating Feature-Feature Similarity Graph.}}
Given an explored feature set $\mathcal{F}$, we create an adjacency matrix to represent the interactions among features, in which each element indicates the edge between two features. For each feature, we use the element values of the corresponding feature within $\mathcal{F}$ as its attributes. Subsequently, we calculate the cosine similarity between every feature pair based on their attributes to determine the weight of the edge, constructing a feature-feature similarity graph. Finally, we convert the similarities adjacency matrix into a 0/1 matrix with the similarity 95th percentile as a cutoff.

\noindent\ul{\emph{Graph Augmentation for Building Contrastive Pairs.}}
We augment each graph with edge perturbation and attribute masking to obtain augmented graphs for building contrastive pairs. The augmented graphs from the same graph form a positive pair, while the augmented graphs from different graphs form negative pairs.
The details of the augmentation strategies are:
1) Edge Perturbation. A random proportion of edges is either added or dropped to perturb the connectivity of the graph. The assumption is that the graph has robustness to the pattern variances of edge connection.
2) Attribute Masking. A random proportion of attributes is masked, prompting the model to use contextual information to recover the obscured information. The assumption is that the missing attributes will not significantly affect the model's predictions.

\noindent\ul{\emph{GNN-based Encoder.}}
The purpose of the GNN-based encoder is to encode the graphs to extract graph-level representation vectors. Since the graphs have different structures (different adjacency matrices), the encoder needs to adapt to various types of graph structures and efficiently handle the learning of node representations. Inspired by~\cite{hamilton2017inductive}, we develop two GNNs with shared weights to encode augmented graphs of different types respectively. 
Here we take one graph as an example to illustrate the encoding process. Given a node $v$ from the graph, we sample its all neighbor nodes, denoted as $\{u_1, u_2, ..., u_p\}$, where $p$ is the number of neighbor nodes. The $\mathbf{h}_v$ and $\mathbf{h}_u$ are the representation of node $v$ and node $u$ respectively. Then we aggregate the representation of all neighbor nodes  by calculating their mean:
$
    \mathbf{h}_u^* = \sum_1^p \mathbf{h}_u / p.
$
We update the representation of node $v$ by
$
    \mathbf{h}_v = ReLU(W \cdot CONCAT(\mathbf{h}_v, \mathbf{h}_u^*)),
$
where $ReLU$ is the activation function and $W$ is the learnable weight matrix.
We iterate this process with multi-step to update the node representation. 
Finally, we calculate the mean of all node presentations in the graph to extract graph-level representation vector for the augmented graph, denoted as: 
$
\mathbf{h} = \sum \mathbf{h}_v/m,
$
where $m$ represents the number of nodes in the graph.
Following the principles advocated in~\cite{chen2020simple}, we use a non-linear projection to project the graph-level representation vectors into a new latent space where the contrastive loss is calculated. In our framework, we use a two-layer Perceptron as a projection to map the $\mathbf{h}$ into a new embedding vector $\mathbf{z}$.

\noindent\ul{\emph{Minimize the Contrastive Loss.}}
Given a $\mathcal{F}_i$ as the $i$-th explored feature set in training feature sets, we convert it into a graph and augment the graph to obtain two augmented graphs of different types. We feed the augmented graphs into the shared GNN respectively, to extract the embedding vectors, which are represented by $\mathbf{z}_{i,1}$ and $\mathbf{z}_{i,2}$ respectively. We calculate the distance between sample pairs using cosine similarity $sim(\mathbf{z}_{i, 1}, \mathbf{z}_{i, 2}) = \mathbf{z}_{i, 1}^T\mathbf{z}_{i, 2}/(\|\mathbf{z}_{i, 1}\|\|\mathbf{z}_{i,2}\|)$, and develop a normalized temperature-scaled cross-entropy loss~\cite{sohn2016improved, oord2018representation} as contrastive loss to maximize the consistency between positive pairs and increase the distance between negative pairs.
\begin{equation}
    \mathcal{L}_{cl} = -\frac{1}{N}\sum_{n=1}^Nlog\frac{exp(sim(\mathbf{z}_{i, 1}, \mathbf{z}_{i, 2})/\tau))}{\sum_{i^*=1, i^* \neq i}^Nexp(sim(\mathbf{z}_{i, 1}, \mathbf{z}_{i^*, 2})/\tau)},
\end{equation}
where $\tau$ denotes the temperature parameter. Finally, we minimize the contrastive loss to optimize the continuous embedding space. 

\subsection{Multi-Objective Fine-tuning}
After pre-training, we conduct multi-objective fine-tuning to update the embedding space. This process aims to assess the efficacy of the embedding space and fully reconstruct the feature cross sequence, thereby facilitating the generation of the optimal transformed feature set.
We reload the pre-trained GNN as an encoder to obtain the embedding. Additionally, we create an evaluator and a decoder for estimating feature set utility and reconstructing feature cross sequence respectively. Then, we identify the best embeddings within the updated embedding space and reconstruct feature cross sequences. Finally, these sequences are employed to transform the original feature space, and the transformed feature set that achieves the highest utility score is presented as the optimal result.

\noindent\textbf{Fine-tuning the Encoder-Decoder-Evaluator Structure.}
We use the training data set $\{\mathcal{F}_i, \mathcal{T}_i, s_i\}_{i=1}^N$ collected in section~\ref{unsupervised_measurement} to fine-tune the encoder-decoder-evaluator structure, where $\mathcal{F}_i$, $\mathcal{T}_i$ and $s_i$ indicate explored feature set, feature cross sequence, and feature set utility of the $i$-th sample, respectively. To simplify notation, we use the notation $(\mathcal{F}, \mathcal{T}, s)$ to represent any set of training data.

\noindent\ul{\emph{The Encoder.}} We input an explored feature set $\mathcal{F}$ into the pre-trained encoder to map the feature set to an embedding vector $\mathbf{z}$.

\noindent\ul{\emph{The Decoder.}} The decoder is to reconstruct $\mathcal{T}$ and built by LSTM~\cite{hochreiter1997long}. Subsequently, we add a softmax layer after the LSTM to estimate the token distribution and infer the feature cross sequence based on the embedding $\mathbf{z}$. Specifically, assuming the token to be decoded is $t_j$, the partially decoded sequence is $[t_1, t_2, ..., t_{j-1}]$, and the length of $\mathcal{T}$ is $k$. Therefore, the probability of the $j$-th token is:
\begin{equation}
    P_\psi(t_j|\mathbf{z}, [t_1, t_2, ..., t_{j-1}]) = \frac{exp(o_j)}{\sum_{k}exp(o)},
\end{equation}
where $o_j$ is the $j$-th output of the softmax layer. We use the negative log-likelihood of generating sequential tokens to minimize the reconstruction loss between the reconstructed tokens and the real one, denoted by:
\begin{equation}
\begin{aligned}
    \mathcal{L}_{rec} =-\sum_{j=1}^k logP_\psi(t_j|\mathbf{z}, [t_1, t_2, ..., t_{j-1}])
\end{aligned}
\end{equation}

\noindent\ul{\emph{The Evaluator.}}
We add a fully connected layer after the encoder for regression prediction, which is given by: $\ddot{s} = \theta(z)$, where $\theta$ is the function notation of the evaluator. Specifically, we use the Mean Squared Error (MSE) to measure the evaluator loss:
\begin{equation}
    \mathcal{L}_{evt} = MSE(\ddot{s}, s),
\end{equation}
where $s$ is the corresponding feature set utility.
Finally, we trade off the two losses to form a joint training loss:
$
    \mathcal{L} = \alpha\mathcal{L}_{evt} + \beta\mathcal{L}_{rec},
$
where $\alpha$ and $\beta$ are the hyperparameters to balance the influence of the two loss functions. We calculate the mean loss of the batch data to readjust the embedding space during the fine-tuning process.

\noindent\textbf{Optimization and Generation of the Optimal Transformed Feature Set.}
After updating the embedding space, we employ a gradient-ascent search method to find better embedding. In the continuous space, a good starting point can accelerate the search process and enhance the downstream performance. Such good points are called initial seeds. Therefore, we first select the top K explored feature sets based on their corresponding feature set utilities as initial seeds to find better embeddings. More specifically, for an initial seed, we obtain an embedding $\mathbf{z}$ through the well-trained encoder. Next, we use a gradient-ascent method to move the embedding with $\eta$ steps along the direction maximizing the feature set utility. Formally, the moving calculation process can be described as:
$
    \mathbf{z}^+ = \mathbf{z} + \eta\frac{\partial\theta}{\partial{\mathbf{z}}},
$
where $z^+$ is the better embedding.
We feed $\mathbf{z}^+$ into the well-trained decoder to generate a feature cross sequence in an autoregressive manner until finding <EOS>. We split the sequence into different segments according to <SEP>, and follow the calculation rule of them to generate the transformed feature set. Finally, the transformed feature set yielding the highest utility score is output as the optimal result. The pseudo-code of the algorithm is released in Appendix~\ref{pseudo-code}.



\section{Experiment}

\setlength{\tabcolsep}{2mm}{
\begin{table*}[tb]
\centering
\small
\caption{Overall Performance. In this table, the best and second-best results are highlighted by \textbf{bold} and \underline{underlined} fonts respectively. We evaluate classification (C) and regression (R) tasks in terms of F1-score and 1-RAE respectively. The higher the value is, the better the quality of the transformed feature set is.}
\vspace{-0.2cm}
\begin{tabular}{@{}c|cccc|ccccccccc|c@{}}
\toprule\toprule
Dataset            & Soruce & C/R & Samples & Features & RDG   & PCA & LDA   &ERG & AFAT  & NFS   & TTG   & GRFG  & DIFER & \model \\ \midrule
Higgs Boston       & UCI & C   & 50000 & 28 & 0.695 &0.538 & 0.513 & \underline{0.702} & 0.697 & 0.691 & 0.699 & \textbf{0.707} & 0.669  & \underline{0.702} \\
Amazon Employee    & Kaggle & C   & 32769 & 9 & 0.932 &0.923 & 0.916 & 0.932 & 0.930  & 0.932 & \underline{0.933} & 0.932 & 0.929  & \textbf{0.936} \\
PimaIndian         & UCI & C   & 768 & 8 & 0.751 &0.736 & 0.729 & 0.735 & 0.732 & \underline{0.791} & 0.745 & 0.754 & 0.760  & \textbf{0.811}  \\
SpectF             & UCI & C   & 267 & 44 & 0.760 &0.709 & 0.665 & 0.788 & 0.760 & \underline{0.792} & 0.760 & 0.776 & 0.766  & \textbf{0.856}  \\
SVMGuide3          & LibSVM & C   & 1243 & 21 & 0.806 &0.676 & 0.635 & \underline{0.818} & 0.794 & 0.786 & 0.798 & 0.812 & 0.773  & \textbf{0.832}  \\
Geman Credit       & UCI & C   & 1000 & 24 & \underline{0.701} &0.679 & 0.596 & 0.662 & 0.639 & 0.649 & 0.644 & 0.667 & 0.656  & \textbf{0.723}  \\
Credit Default     & UCI & C   & 30000 & 24 & 0.802 &0.768 & 0.743 & 0.803 & 0.804 & 0.801 & 0.798 & \underline{0.806} & 0.796  & \textbf{0.809}  \\
Messidor\_features & UCI & C   & 1151 & 19 & 0.627 &0.672  & 0.463 & \underline{0.692} & 0.657 & 0.650  & 0.655 & 0.680 & 0.660  & \textbf{0.737} \\
Wine Quality Red   & UCI & C   & 999 & 12 & \underline{0.496} &0.468 & 0.401 & 0.485 & 0.480 & 0.451 & 0.467 & 0.454 & 0.476  & \textbf{0.522}  \\
Wine Quality White & UCI & C   & 4898 & 12 & 0.524 &0.469 & 0.438 & \underline{0.527} & 0.516 & 0.525 & 0.521 & 0.518 & 0.507  & \textbf{0.554}  \\
Spambase           & UCI & C   & 4601 & 57 & 0.906 &0.815 & 0.889 & 0.917 & 0.912 & \underline{0.925} & 0.919 & 0.922 & 0.912  & \textbf{0.932}  \\
Ap-omentum-ovary   & Openml & C   & 275 & 10936 & 0.820 &0.718 & 0.716 & \underline{0.849} & 0.813 & 0.830 & 0.830 & 0.830 & 0.833  & \textbf{0.865}  \\
Lymphography       & UCI & C   & 148 & 18 & 0.108 &0.170 & 0.144 & \underline{0.202} & 0.149 & 0.166 & 0.148 & 0.136 & 0.150  & \textbf{0.298}  \\
Ionosphere         & UCI & C   & 351 & 34 & 0.942 &0.928 & 0.743 & \underline{0.957} & 0.942 & 0.943 & 0.932 & 0.956 & 0.905  & \textbf{0.971}  \\ \midrule
Housing Boston     & UCI & R   & 506 & 13 & \underline{0.434} &0.122 & 0.020 & 0.410 & 0.401 & 0.428 & 0.396 & 0.409 & 0.381  & \textbf{0.478}  \\
Airfoil            & UCI & R   & 1503 & 5 & 0.519 &0.514 & 0.207 & 0.519 & 0.507 & 0.520 & 0.500 & 0.521 & \underline{0.528}  & \textbf{0.578} \\
Openml\_586        & Openml & R   & 1000 & 25 & 0.568 &0.221 & 0.110 & 0.624 & 0.553 & 0.542 & 0.544 & \underline{0.646} & 0.482  & \textbf{0.700}  \\
Openml\_589        & Openml & R   & 1000 & 50 & 0.509 &0.224 & 0.011 & 0.610 & 0.508 & 0.503 & 0.504 & \textbf{0.627} & 0.463  & \underline{0.625}  \\
Openml\_607        & Openml & R   & 1000 & 50 & 0.521 &0.104 & 0.107 & 0.549 & 0.509 & 0.518 & 0.522 & \textbf{0.555} & 0.476  & \underline{0.550}  \\
Openml\_616        & Openml & R   & 500 & 50 & 0.070  &0.057 & 0.024 & 0.193 & 0.162 & 0.147 & 0.156 & \underline{0.372} & 0.076  & \textbf{0.536}  \\
Openml\_618        & Openml & R   & 1000 & 50 & 0.472 &0.078 & 0.052 & 0.561 & 0.473 & 0.467 & 0.467 & \underline{0.577} & 0.408  & \textbf{0.616}  \\
Openml\_620        & Openml & R   & 1000 & 25 & 0.511 &0.088 & 0.029 & \underline{0.546} & 0.521 & 0.519 & 0.512 & 0.530 & 0.442  & \textbf{0.584}  \\
Openml\_637        & Openml & R   & 500 & 50 & 0.136 &0.072 & 0.043 & 0.064 & 0.161 & 0.146 & 0.144 & \textbf{0.307} & 0.072  & \underline{0.278}  \\ \bottomrule\bottomrule
\end{tabular}
\vspace{-0.3cm}
\label{exp:overall_performance}
\end{table*}}

\begin{figure*}[htp]
        \centering
        \subfigure[SpectF]{\label{exp:spect}\includegraphics[width=0.163\textwidth]{{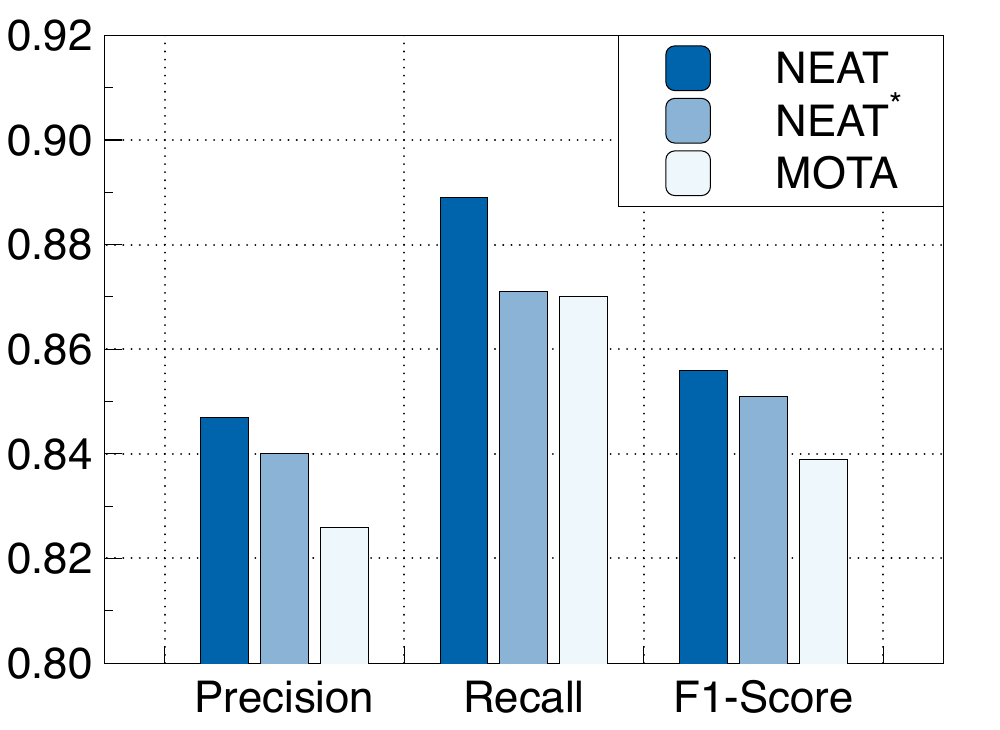}}}
        \subfigure[SVMGuide3]{\label{exp:svmguuide}\includegraphics[width=0.163\textwidth]{{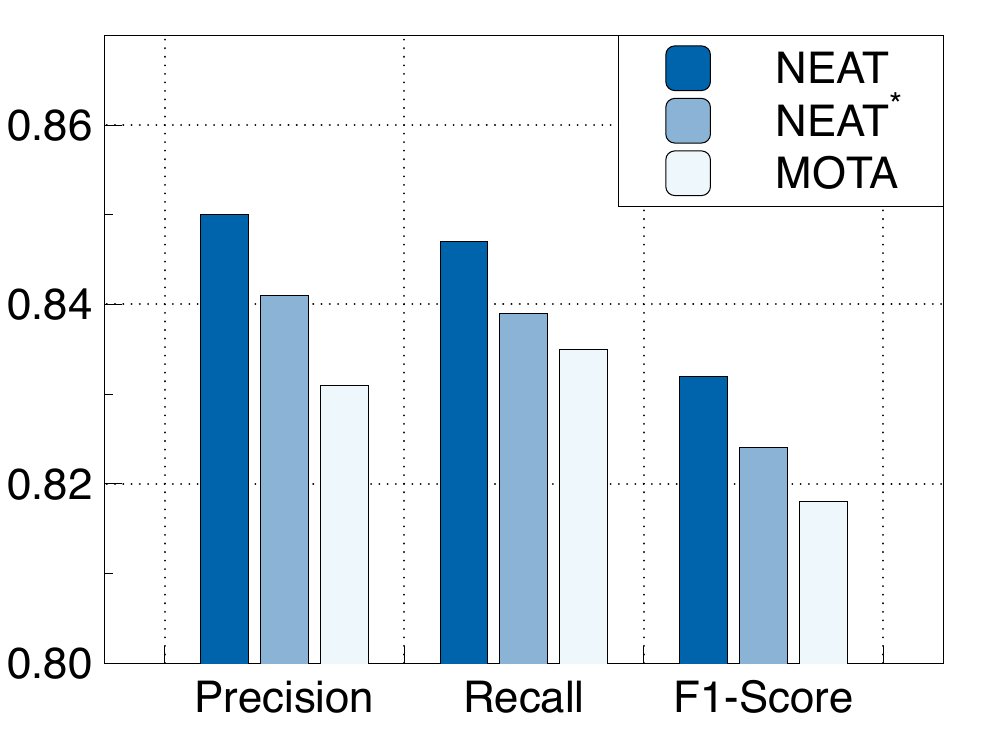}}}
        \subfigure[German Credit]{\label{exp:german}\includegraphics[width=0.163\textwidth]{{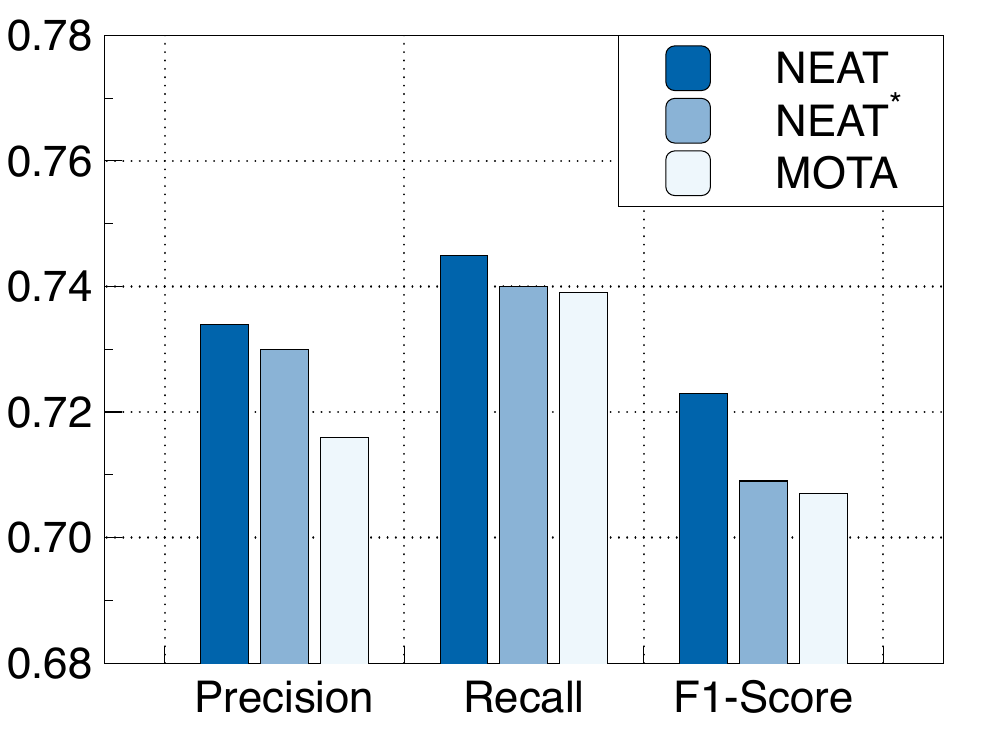}}}
        \subfigure[Openml\_616]{\label{exp:openml616}\includegraphics[width=0.163\textwidth]{{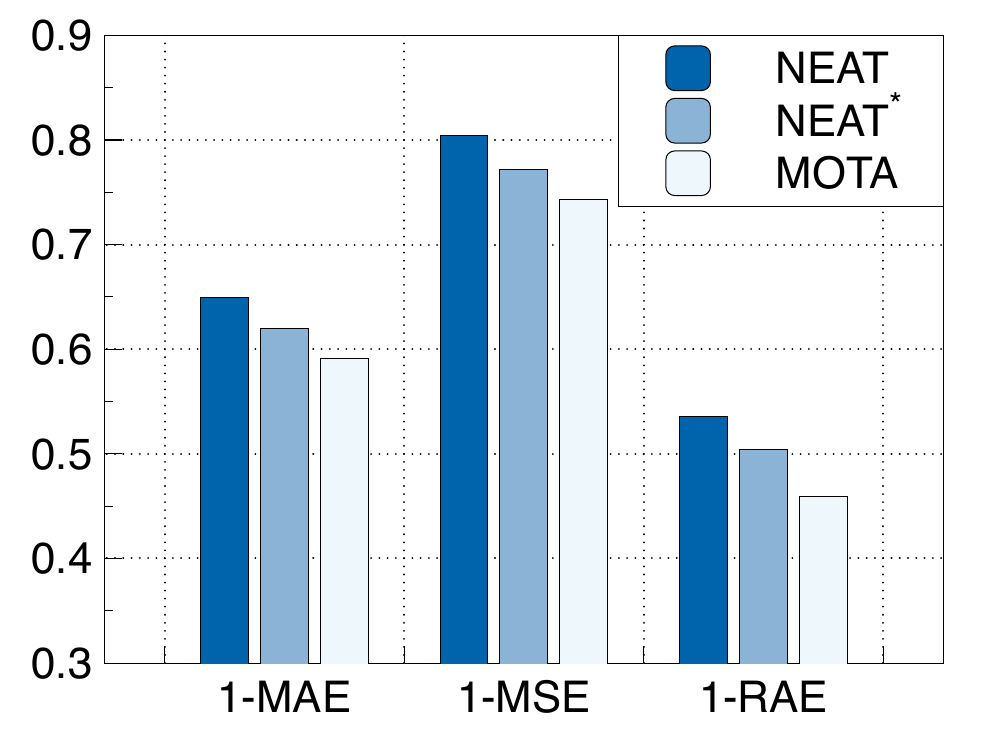}}}
        \subfigure[Openml\_618]{\label{exp:openml618}\includegraphics[width=0.163\textwidth]{{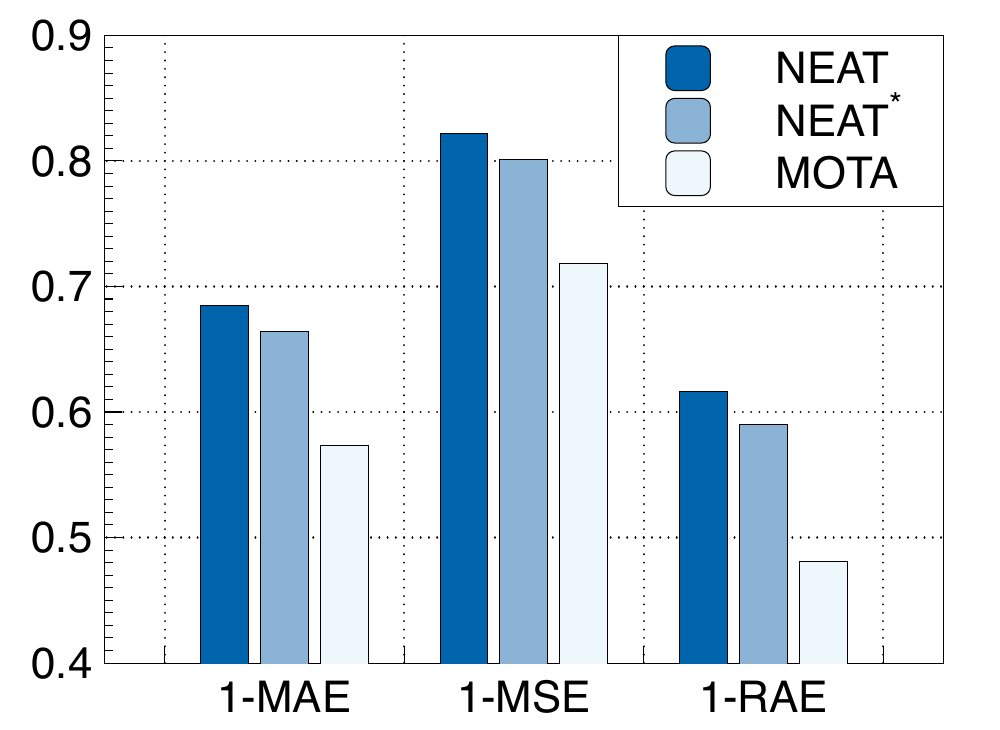}}}
        \subfigure[Openml\_620]{\label{exp:openml620}\includegraphics[width=0.163\textwidth]{{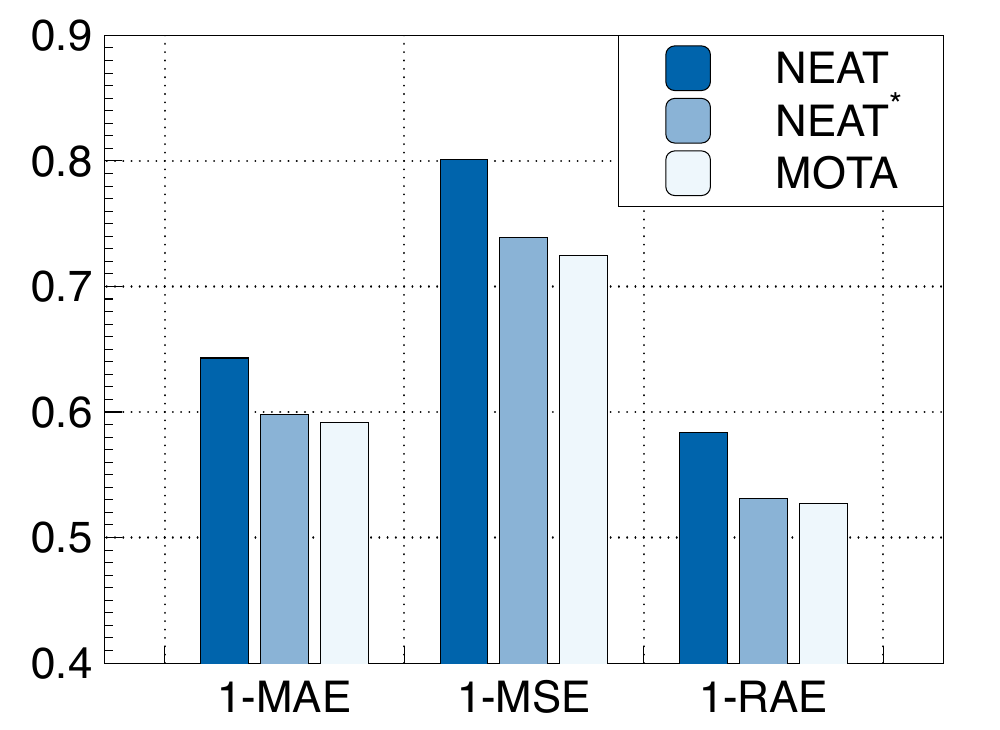}}}
        \vspace{-0.4cm}
        \caption{The impact of graph contrastive pre-training.}
        \vspace{-0.3cm}
        \label{fig:CL-exp}
\end{figure*}

\begin{figure*}[htp]
        \centering
        \subfigure[SpectF]{\label{exp:spect}\includegraphics[width=0.163\textwidth]{{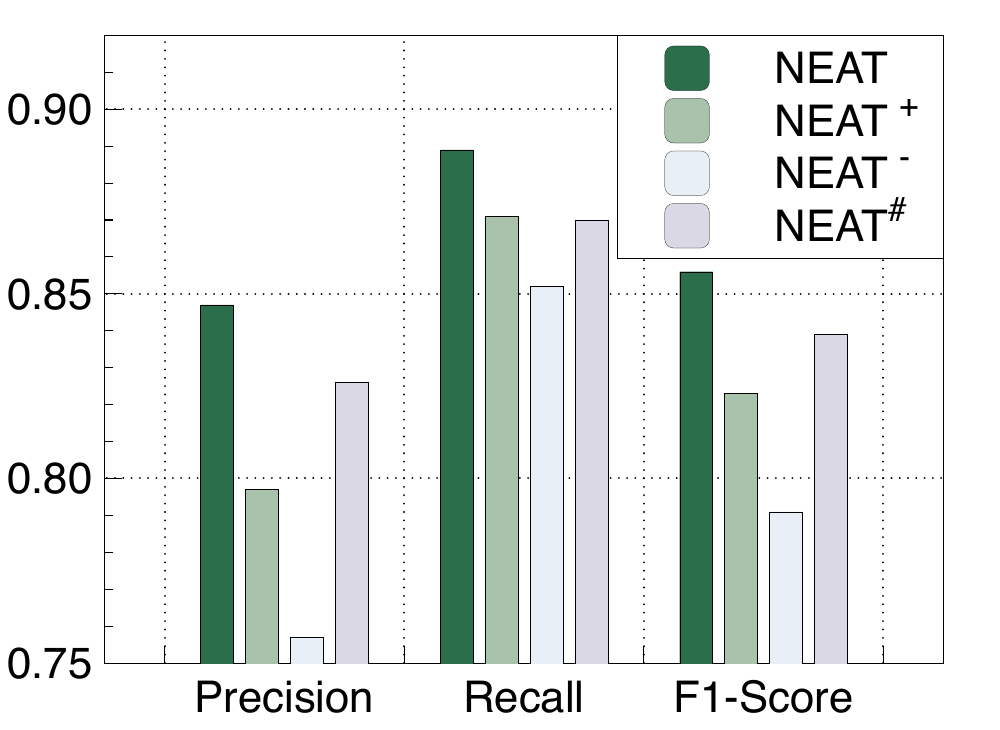}}}
        \subfigure[SVMGuide3]{\label{exp:svmguuide}\includegraphics[width=0.163\textwidth]{{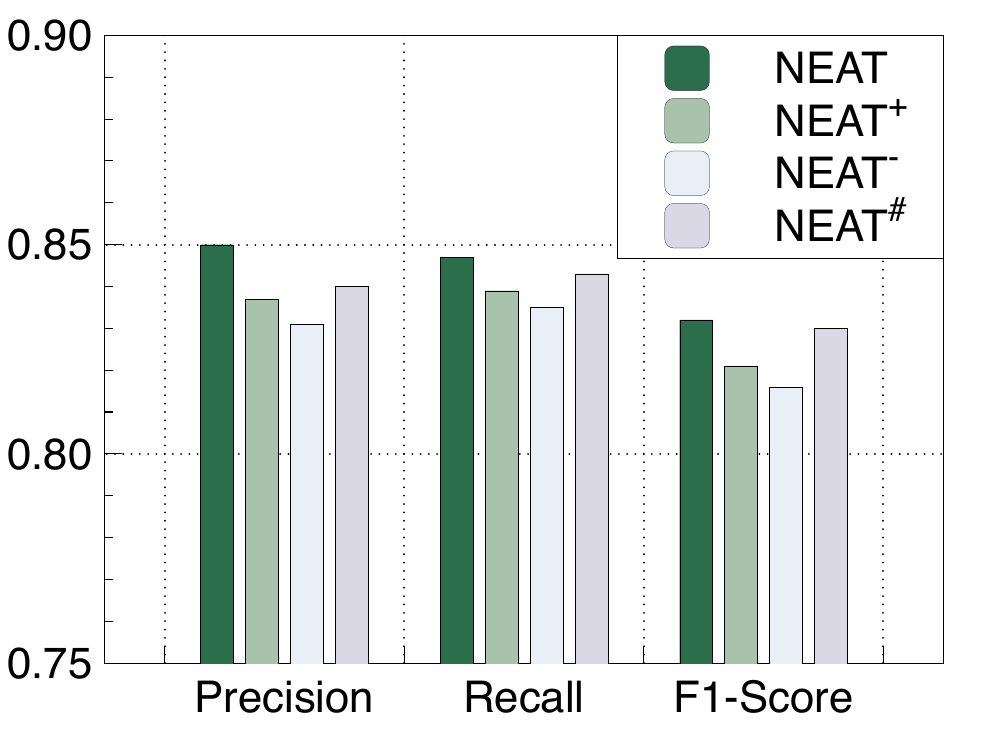}}}
        \subfigure[German Credit]{\label{exp:german}\includegraphics[width=0.163\textwidth]{{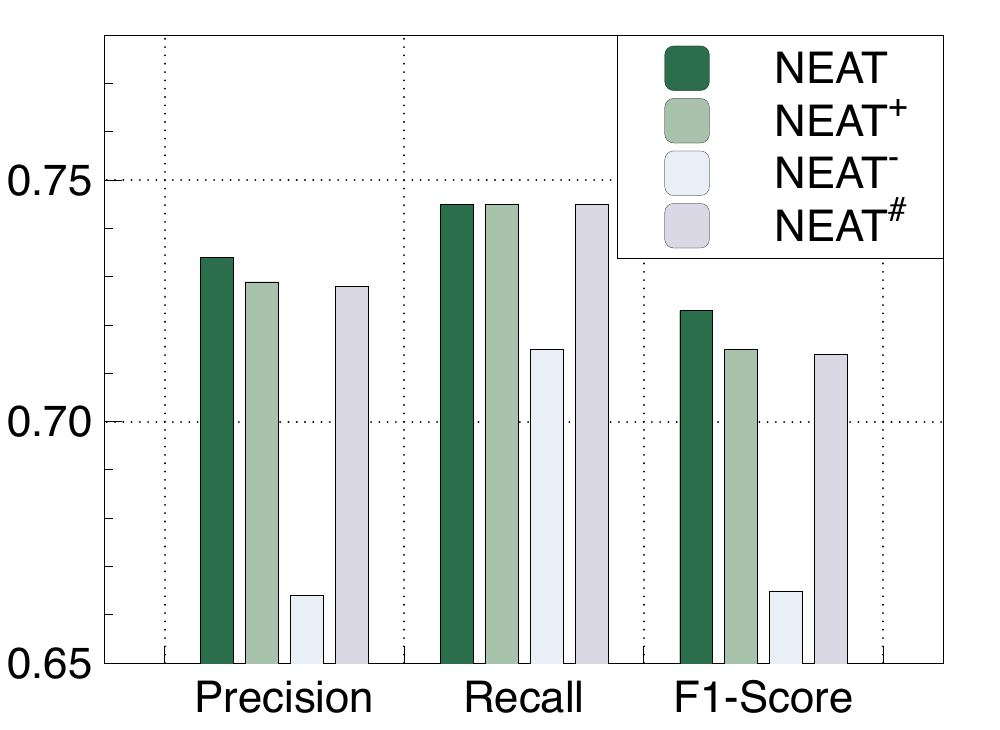}}}
        \subfigure[Openml\_616]{\label{exp:openml616}\includegraphics[width=0.163\textwidth]{{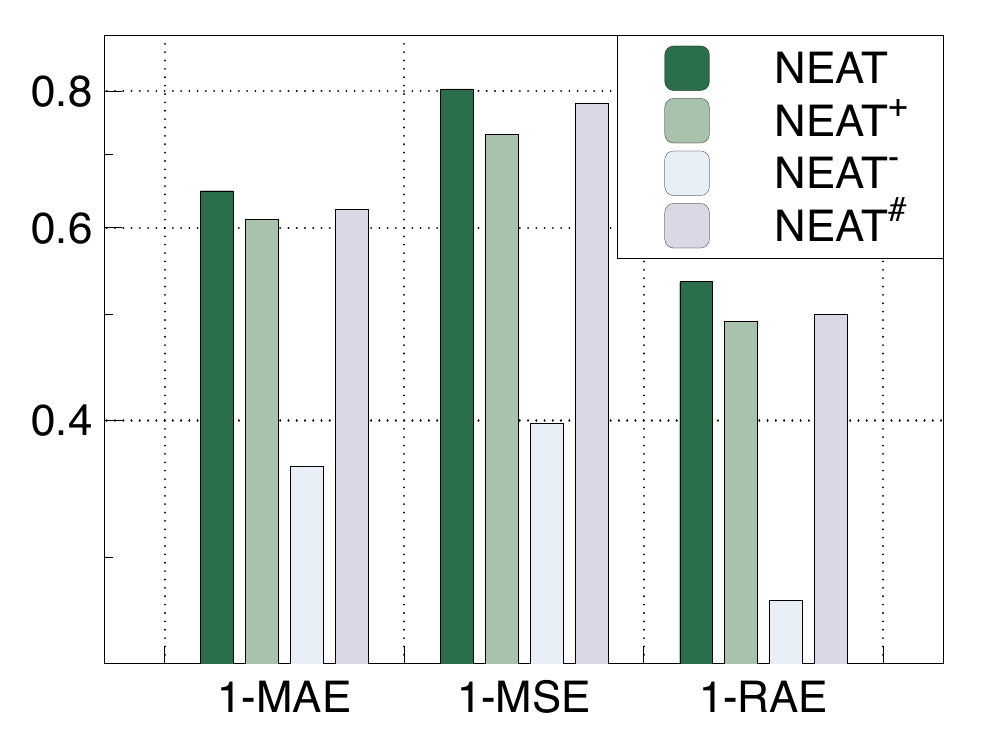}}}
        \subfigure[Openml\_618]{\label{exp:openml618}\includegraphics[width=0.163\textwidth]{{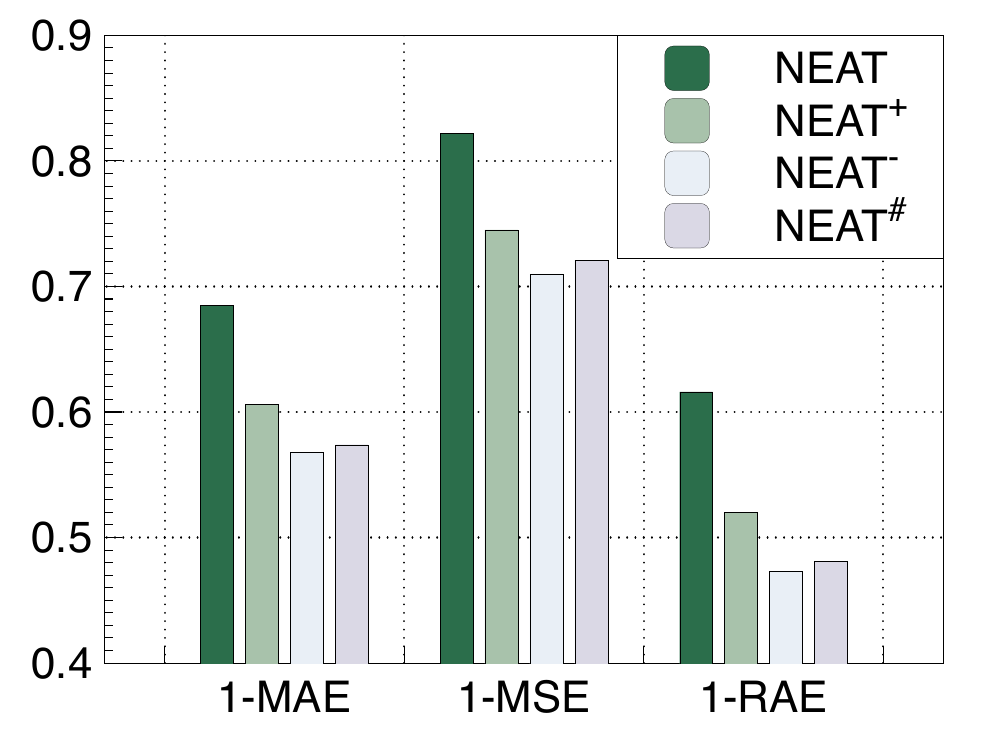}}}
        \subfigure[Openml\_620]{\label{exp:openml620}\includegraphics[width=0.163\textwidth]{{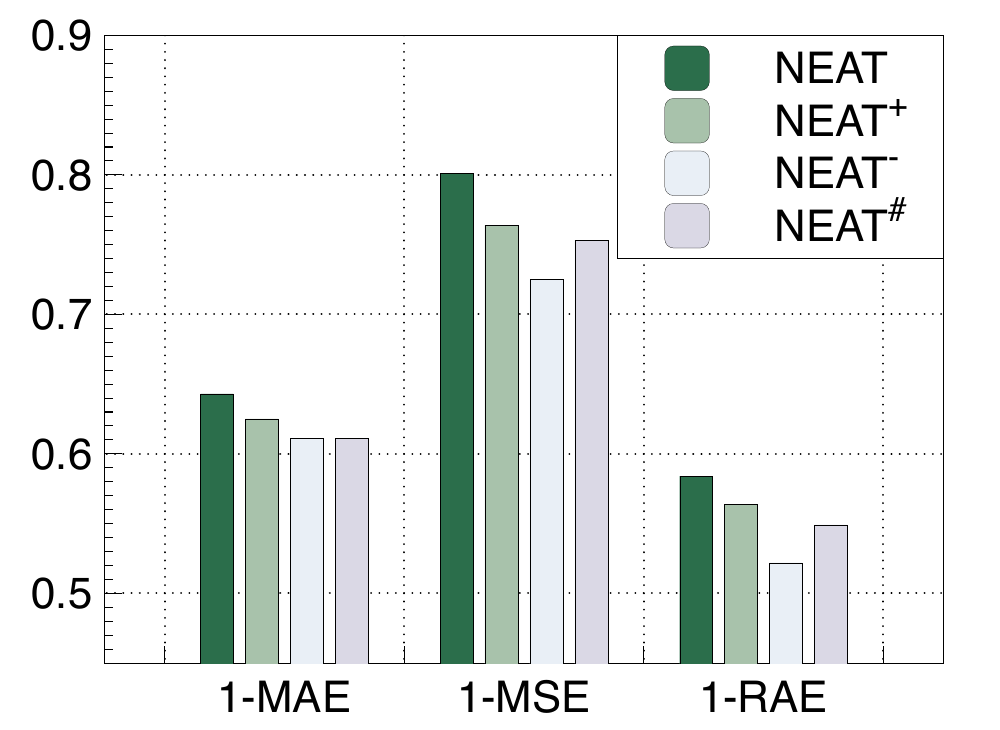}}}
        \vspace{-0.4cm}
        \caption{The impact of unsupervised feature set utility measurement, RL-based data collector, and initial seeds.}
        \vspace{-0.3cm}
        \label{fig:data-init}
\end{figure*}

\subsection{Experimental Setup}
\noindent\textbf{Dataset Descriptions.}
We utilize 23 publicly available datasets from UCI, LibSVM, Kaggle, and OpenML to conduct experiments. These datasets comprise 14 classification tasks (C) and 9 regression tasks (R). Table~\ref{exp:overall_performance} shows the statistics of these datasets.

\noindent\textbf{Evaluation Metrics.}
To alleviate the variance of the downstream ML model for fair comparisons, we employ random forests (RF) to assess the quality of transformed feature set. 
For classification tasks, we use F1-score, Precision, and Recall as evaluation metrics. 
For regression tasks, we use 1 - Relative Absolute Error (1-RAE), 1 - Mean Average Error (1-MAE), and 1 - Mean Square Error (1-MSE) as evaluation metrics.
In both cases, higher values of the evaluation metrics indicate a higher quality of the transformed feature set.

\noindent\textbf{Baseline Algorithms and Model Variants of \model.}
We compare our method with 9 widely-used feature transformation methods:
(A) Unsupervised Methods:
1) \textbf{RDG} randomly generates feature-operation-feature transformation records to create a new feature space;
2) \textbf{PCA}~\cite{uddin2021pca} uses linear feature correlation to generate new features;
3) \textbf{LDA}~\cite{LDA} is a matrix factorization-based method that obtains decomposed latent representation as the generated feature space; 
(B) Supervised Methods: 
3) \textbf{ERG} first applies a set of operations to each feature to expand the feature space, and then selects key features from it to form a new feature space; 
4) \textbf{AFAT}~\cite{expansion-reduction2} is an enhanced version of ERG, repeatedly generating new features and using a multi-step feature selection method to select more informative features for creating a new feature space; 
5) \textbf{NFS}~\cite{automl1} embeds the decision-making process of feature transformation into a policy network and utilizes reinforcement learning (RL) to optimize the entire feature transformation process; 
6) \textbf{TTG}~\cite{envolution2} represents the feature transformation process as a graph and implements an RL-based discrete search method to find the optimal feature set; 
7) \textbf{GRFG}~\cite{envolution1} uses three collaborative reinforcement intelligent agents for feature generation and employs feature grouping strategies to accelerate agent learning; 
8) \textbf{DIFER}~\cite{automl2} embeds randomly generated feature cross sequences and then uses a greedy search to find the separated transformed feature.

Additionally,
to evaluate the necessity of each technical component of \model, we implement four model variants:
1) \textbf{\model$^*$} removes the pre-training stage and directly optimizes the encoder-evaluator-decoder structure for creating the feature transformation embedding space. 
2) \textbf{\model$^+$} utilizes the training data collected by randomly generating explored feature sets to create the feature transformation embedding space.
3) \textbf{\model$^-$} replaces the mean discounted cumulative gain with feature redundancy for creating the feature transformation embedding space.
4) \textbf{\model$^\#$} replaces the initial seeds with worst-ranked transformation sequences' embeddings based on the utility value to identify better feature spaces.

\noindent\textbf{Hyperparameters and Reproducibility.}
1) Feature Transformation Knowledge Acquisition: We use the reinforcement data collector to collect explored feature set-feature utility score pairs.
The collector explores 512 episodes, and each episode includes 10 steps.
2) Feature-Feature Similarity Graph Construction: The threshold for creating an edge between two features is the 95th percentile of all similarity values.
3) Graph Contrastive Pre-training: We map the attribute of each node to a 64-dimensional embedding, and use a 2-layer GNN network and a 2-layer projection head to integrate such information.
For graph augmentation, we randomly perturbed 20\% of the edges or masked 20\% of the attributes to get diversified graphs.
During the optimization, the batch size is 1024, the pre-training epochs are 100, the learning rate is 0.001, and the value of $\tau$ is set to 0.5.
4) Multi-objective Fine-tuning: The decoder is a 1-layer LSTM network, which reconstructs a feature cross sequence. 
The evaluator is a 2-layer feed-forward network, in which the dimension of each layer is 200. 
During optimization, $\alpha$ and $\beta$ are 10 and 0.1 respectively, the batch size is 1024, the fine-tuning epochs are 500, and the learning rate is 0.001. 

\noindent\textbf{Environmental Settings.}
All experiments are conducted on the Ubuntu 22.04.3 LTS operating system, Intel(R) Core(TM) i9-13900KF CPU@ 3GHz, and 1 way RTX 4090 and 32GB of RAM, with the framework of Python 3.11.4 and PyTorch 2.0.1.

\vspace{-0.1cm}
\subsection{Experimental Results}

\noindent\textbf{Overall performance.} 
In this experiment, we compare the performance of \model\ and baseline models for feature transformation in terms of F1-score and 1-RAE. Table~\ref{exp:overall_performance} shows the comparison results. We can find that in most cases, the \model\ performs the best, and in the other four cases, it ranks second-best. But \model\ performs the best across all unsupervised approaches.
The underlying driver for this observation is that \model\ can accurately capture the internal principle of the features for identifying the optimal feature space through the integration of unsupervised measurement, pre-training, and fine-tuning stages.
Moreover, another interesting observation is that \model\ exhibits a higher degree of robustness and effectiveness in its transformation performance when applied to classification tasks as compared to regression tasks.
A potential reason for this observation is that \model\ effectively captures the intricate relationships within the feature space, generating non-linear features to enhance the distinguishability of the feature space.
Such transformations have more influence on classification tasks than regression tasks.
In conclusion, this experiment shows the effectiveness of \model\ in feature transformation, underscoring the great potential of generative AI in this domain.

\begin{figure}[]
        \centering
        \subfigure[GPU Memory Utilization]{\label{exp:graphic-memory}\includegraphics[width=0.235\textwidth]{{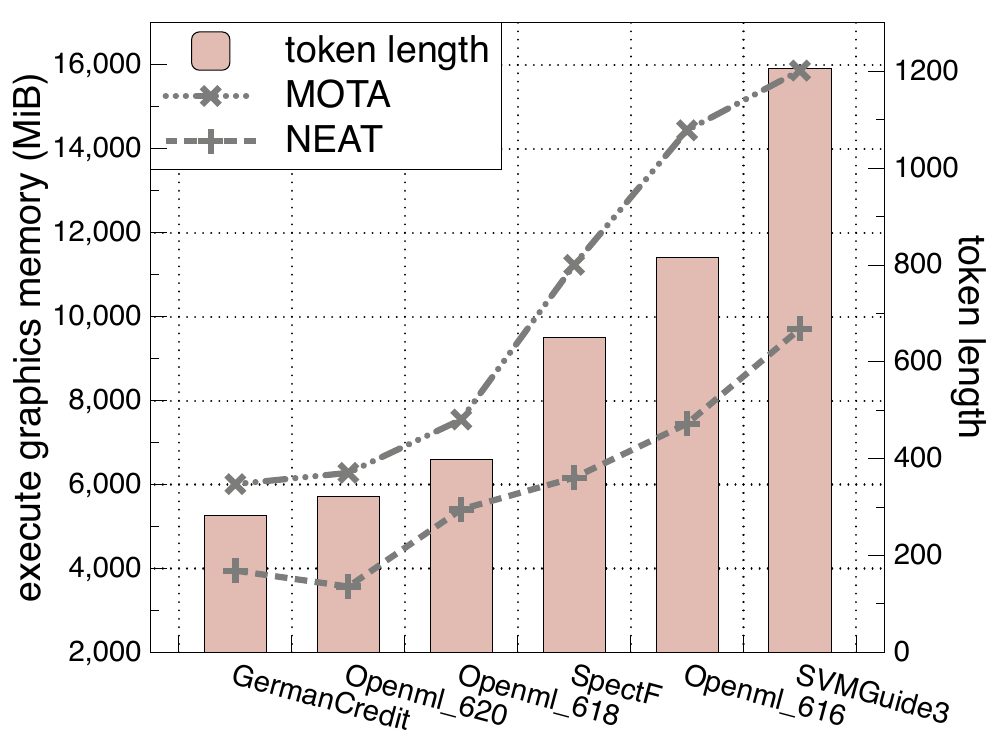}}}
        \subfigure[Convergence Speed]{\label{exp:converge-speed}\includegraphics[width=0.235\textwidth]{{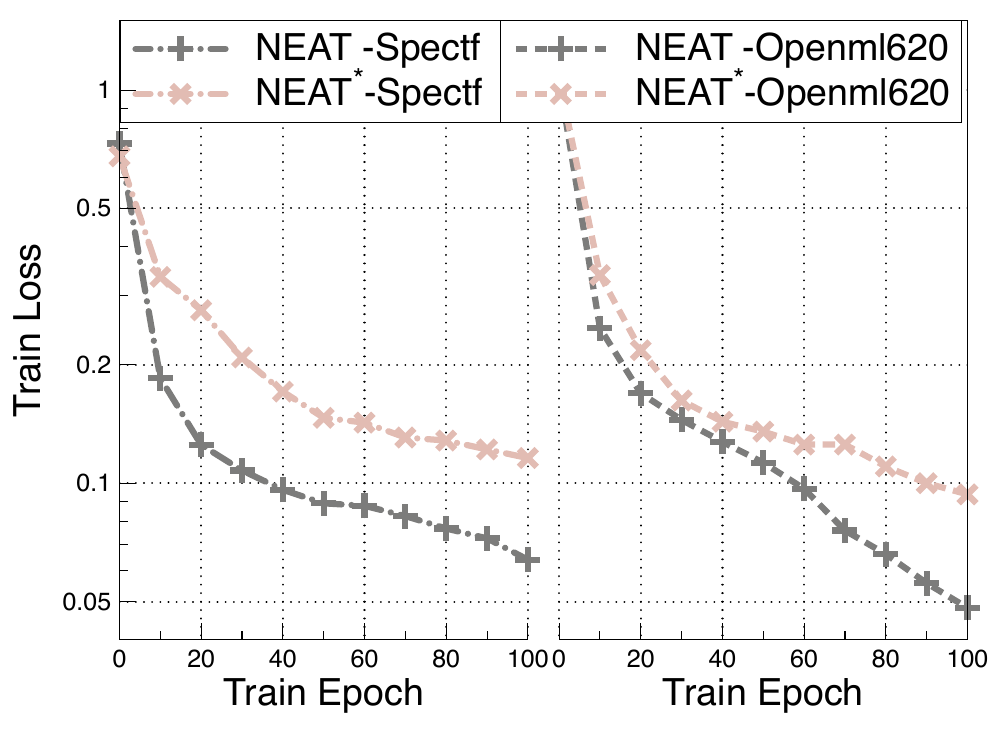}}}
        \vspace{-0.4cm}
        \caption{(a) GPU memory utilization comparison between {\model} and MOTA in different datasets. (b) Convergence speed comparison between {\model} and {\model$^*$}}
        \vspace{-0.4cm}
        \label{gpu_con}
\end{figure}
\begin{figure}[]
        \centering
        \subfigure[Data collection time and model size]{\label{exp:collect-time}\includegraphics[width=0.235\textwidth]{{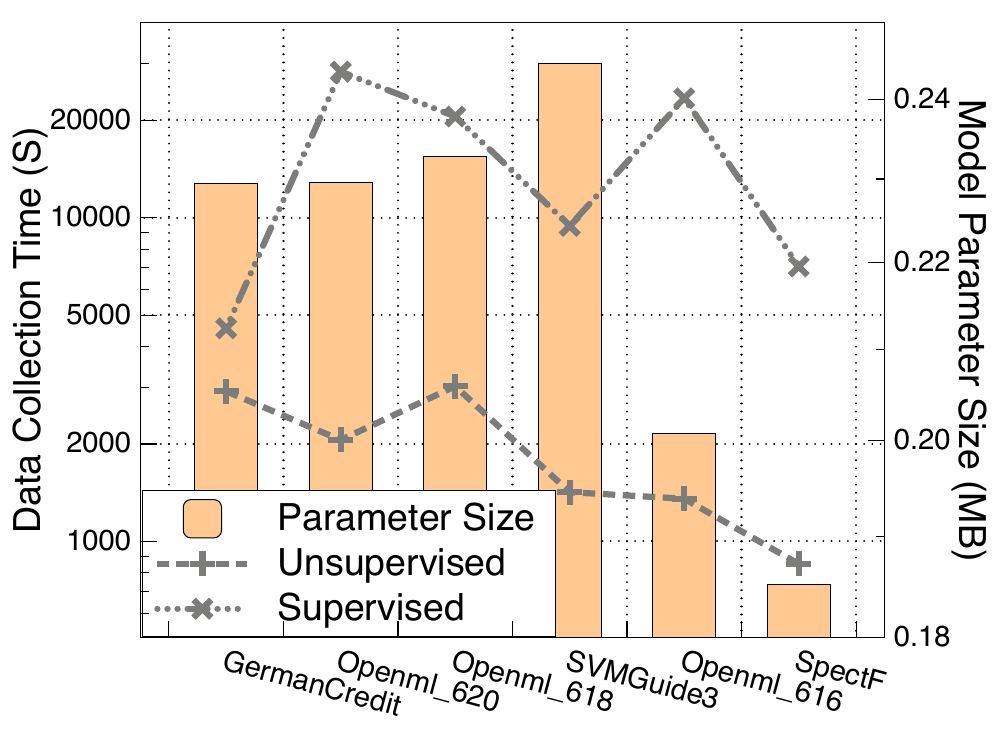}}}
        \subfigure[Training and inference time]{\label{exp:train-time}\includegraphics[width=0.235\textwidth]{{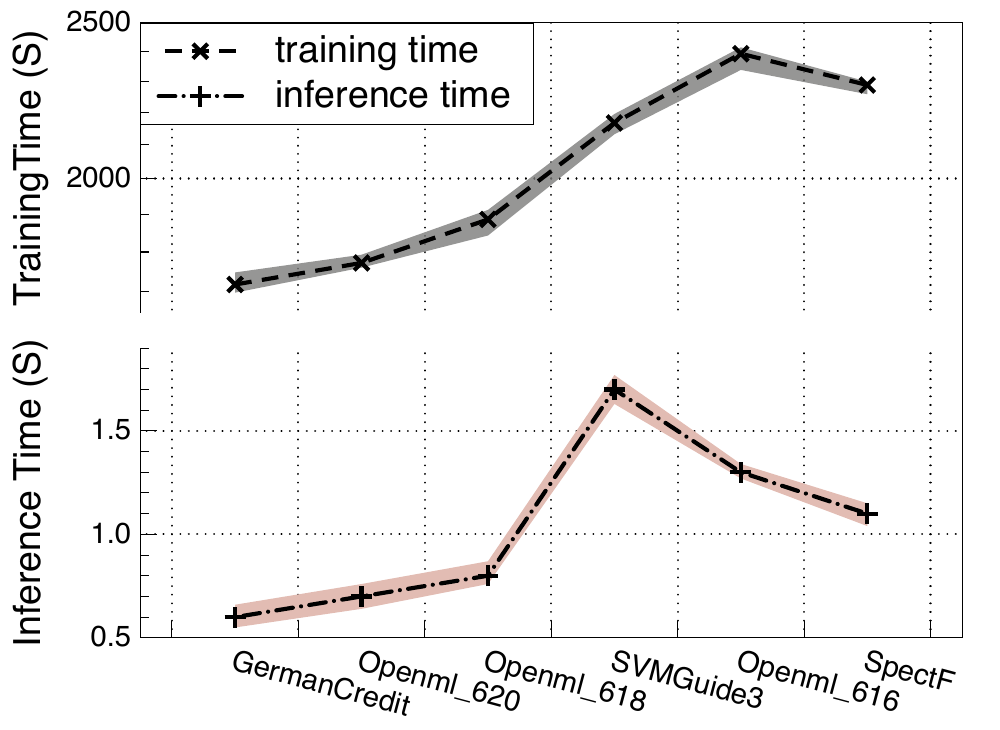}}}
        \vspace{-0.4cm}
        \caption{(a) Data collection time comparison in terms of unsupervised and supervised collectors. Model parameter size check. (b) Model training and inference time check.}
        \vspace{-0.4cm}
        \label{graphic_memory}
\end{figure}

\noindent\textbf{The impact of the graph contrastive learning based pretraining.}
One of the most important novelties of \model\ is to involve graph contrastive learning to pre-train the transformation embedding space. To analyze the influence of this component, we develop two model variants:
1) \model$^*$;
2) MOTA~\cite{wang2023reinforcement}
is similar to \model, but it lacks the pre-training stage and employs an LSTM encoder. 
To ensure fair comparisons, we use our defined feature utility score as feature space measurement in both MOTA and \model.
Figure~\ref{fig:CL-exp} shows the comparison results on both classification and regression tasks.
We can find that \model\ outperforms \model$^*$\ and MOTA in all cases. The underlying driver is that the pre-training stage initializes an effective transformation embedding space by capturing intricate relationships within the feature space. 
This initialization integrates with multi-objective fine-tuning to enhance the identification of the optimal feature space.
Another interesting observation is that \model$^*$\ beats MOTA in most cases. 
A potential reason is that the GNN-based encoder is adept at grasping feature-feature interactions and correlations compared with LSTM to produce an improved transformation embedding space for better reconstruction.
In summary, this experiment demonstrates the necessity of the pre-training stage and the importance of the GNN-based encoder.

\noindent\textbf{The impact of unsupervised feature set utility measurement, RL-based data collector, and initial seeds.}
In \model, to create an effective transformation embedding space, we adopt three strategies: 
1) Using an RL-based data collector~\cite{envolution1} to automatically collect high-quality transformation records.
2) Developing an unsupervised measurement to evaluate the transformation records' utility.
3) Employing top-ranked transformation sequences, determined by their feature utility score, as initial seeds to search for enhanced transformation sequences.
We develop three model variants to study their impact:
\model$^+$, \model$^-$, and \model$^\#$.
Figure~\ref{fig:data-init} shows the comparison results.
We find that \model\ surpasses \model$^+$, \model$^-$ and and \model$^\#$ across all datasets.
There are three underlying drivers: 
a) The RL-based collector can mimic human intelligence in feature learning to produce high-quality transformation records.
These records help create an informative and distinguishable embedding space to identify the optimal feature space.
b) The mean discounted cumulative gain, which considers the feature value consistency preservation, can better evaluate the feature set utility compared to feature redundancy.
c) \model\ converts feature generation as a continuous optimization task, in which a good initialization can enhance the search for the optimal transformation sequence.

\noindent\textbf{Analysis of GPU utilization and convergence speed for \model.}
To further analyze the GPU utilization and convergence speed of \model, we compare it with both MOAT and \model$^*$.
Figure~\ref{gpu_con} shows the comparison results.
From Figure~\ref{exp:graphic-memory}, we find that \model\ significantly saves the required GPU memory compared with MOTA.
A potential reason is that MOTA employs LSTM as the encoder, with the increase in feature space dimension, the required GPU memory will increase correspondingly.
Instead, \model\ organizes the original feature space as a feature-feature attributed graph and utilizes GNNs to aggregate information into embedding.
This learning approach is more memory-efficient.
From Figure~\ref{exp:converge-speed}, we observe that \model\ converges faster and achieves lower loss compared with \model$^*$. 
A possible reason is that the pre-training stage preserves complex feature-feature relationships within the transformation embedding space.
This accelerates the learning process of the fine-tuning stage to achieve model convergence.
In summary, this experiment shows the efficient GPU memory usage and fast convergence of \model.

\noindent\textbf{Robustness check.}
To study the robustness of \model, we compare its transformation performance and other baselines across various downstream ML models.
We replace the Random Forest (RF) with XGBoost (XGB), Support Vector Machine (SVM), K-Nearest Neighborhood (KNN), Decision Tree (DT), and LASSO respectively. Table~\ref{exp:robustness_check} shows the F1-score comparison results for the German Credit dataset.
We can find that \model\  consistently exhibits superior performance in most cases.
A potential reason is that the unsupervised feature utility measurement of \model\ leads it to grasp more general and intrinsic characteristics of the feature space.
Such information is captured independently of the target labels, thereby enhancing the robustness and generalization capabilities of \model.
In summary, this experiment demonstrates the robustness of \model, which maintains consistent performance across various ML models.

\noindent\textbf{Time and space complexity analysis.}
To check the time and space complexity of \model, we select 6 datasets to compare data collection time, model parameter size, training time, and inference time. Figure~\ref{exp:collect-time} shows the comparison results in terms of the time cost of the data collector and model parameter size.
We find that the data collector using our defined unsupervised feature utility measurement requires much less collection time compared with the supervised collector using the target label. 
This observation shows the computational efficiency of our unsupervised measurement.
A possible reason is that the supervised collector has to spend much time retraining the downstream ML model from scratch at each iteration.
Moreover, we notice that \model\ can maintain a very small parameter size across all datasets.
A potential reason is that our sampling-based GNN encoder leads \model\ to focus on key information within the feature space instead of the global situation, significantly increasing the scalability of \model.
Furthermore, Figure~\ref{exp:train-time} shows the comparison results in terms of training and inference time costs. 
We find that once \model\ converges, it can quickly infer and identify the optimal feature space. 
The exceptional inference time of \model\ underscores its great practical potential. We further compare the time cost of {\model} with various baselines in Appendix~\ref{time-complexity-baselines}, and analyze the time cost of the different components within {\model} in Appendix~\ref{app-diff}.
\setlength{\tabcolsep}{3mm}{
\begin{table}[t]
\centering
\small
\caption{Robustness check. Test the transformed feature set on various downstream ML models.}
\vspace{-0.3cm}
\begin{tabular}{@{}c|cccccc@{}}
\toprule
       & RF    & XGB   & SVM   & KNN   & DT    & LASSO \\ \midrule
RDG    & \underline{0.701} & 0.690  & 0.694 & \underline{0.660}  & 0.652 & 0.666 \\
PCA    & 0.679 & 0.641 & 0.657 & 0.641 & 0.676 & 0.659 \\     
LDA    & 0.596 & 0.629 & 0.589 & 0.622 & 0.604 & 0.589 \\
ERG    & 0.662 & 0.669 & 0.599 & 0.619 & 0.630 & \textbf{0.684} \\
AFAT   & 0.639 & 0.688 & 0.639 & 0.643 & \underline{0.691} & 0.647 \\
NFS    & 0.649 & 0.690 & \underline{0.695} & 0.635 & 0.663 & 0.672 \\
TTG    & 0.644 & 0.690 & \underline{0.695} & \underline{0.660} & 0.663 & 0.666 \\
GRFG   & 0.667 & 0.656 & 0.685 & 0.649 & 0.687 & 0.676 \\
DIFER & 0.656 & \underline{0.696} & 0.617 & 0.658 & 0.683 & 0.672 \\ \midrule
\model  & \textbf{0.723} & \textbf{0.707} & \textbf{0.703} & \textbf{0.661} & \textbf{0.692} & \underline{0.676} \\ \bottomrule
\end{tabular}
\vspace{-0.4cm}
\label{exp:robustness_check}
\end{table}}
\begin{figure}[t]
        \centering
        \subfigure[Originial]{\label{exp:original}\includegraphics[width=0.235\textwidth]{{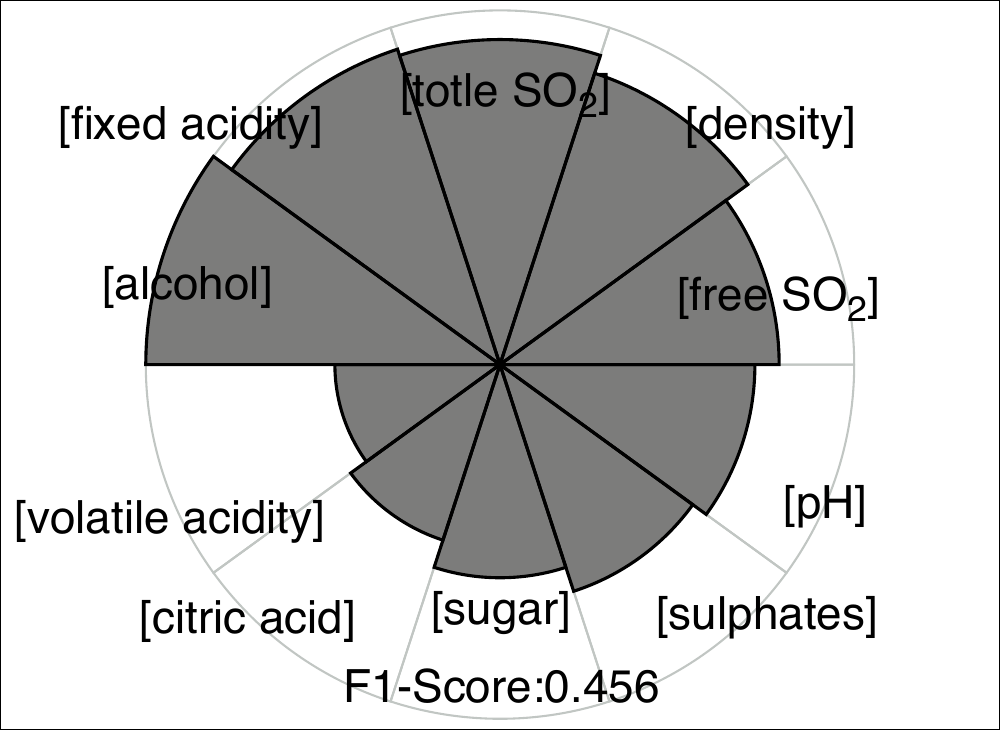}}}
        \subfigure[Transformation]{\label{exp:transformed}\includegraphics[width=0.235\textwidth]{{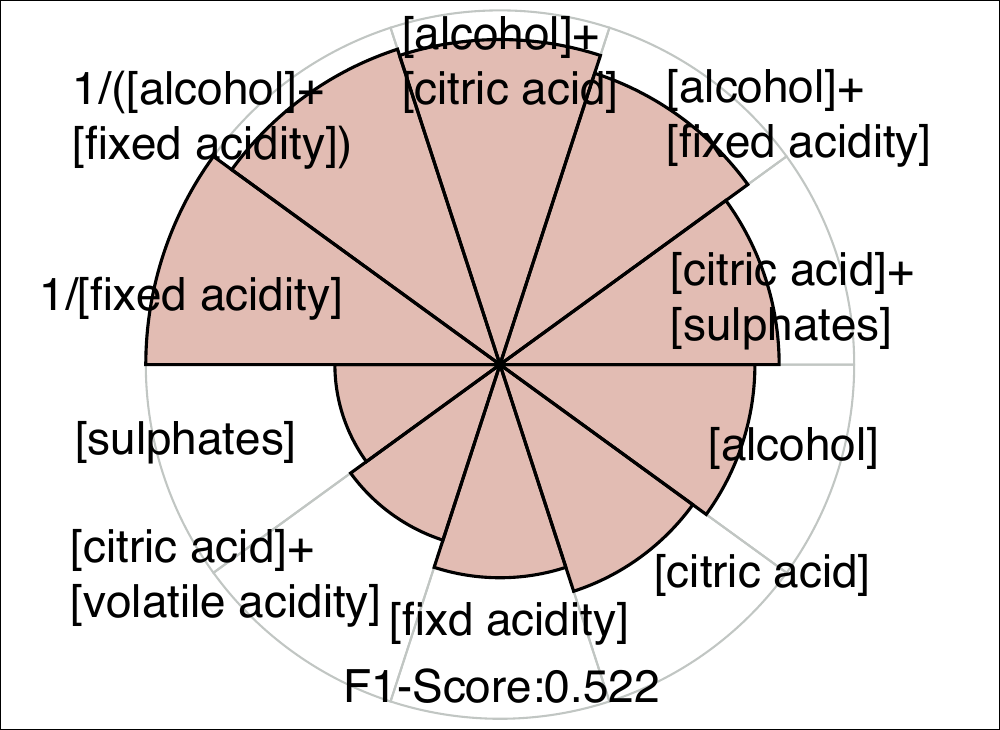}}}
        \vspace{-0.4cm}
        \caption{Case study. Comparison of traceability on the original feature space and transformed feature space.}
        \vspace{-0.4cm}
        \label{case-explanation}
\end{figure}

\noindent\textbf{Case study.}
We employ our proposed unsupervised feature utility measurement (See Step 3 in Section~\ref{unsupervised_measurement}) to calculate the feature importance of the original and transformed feature space respectively using the Wine Quality Red dataset. 
This task aims to predict the wine quality using the collected features.
Figure~\ref{case-explanation} shows the comparison results in terms of the top 10 important features. 
In each pie chart, the importance of a feature is indicated by the area of the corresponding slice.
The text labeling each slice of the pie chart denotes the name of the associated feature.
We observe that in the transformed feature space, 60\% of the top 10 features are generated by \model.
This contributes to 14.5\% improvements in wine quality prediction. 
Such an observation indicates that  \model\ comprehends the complex inherent relationships of the feature set, thereby generating a better feature space.
Furthermore, we notice that 'alcohol' and 'fixed acidity' are often employed to generate new informative features, such as '1/[fixed acidity]' and '1/([alcohol]+[fixed acidity])'.
The observation aligns with known wine knowledge: alcohol enhances taste and body, while fixed acidity influences tartness and stability, collectively determining wine quality.
\vspace{-0.1cm}
\section{Related Work}\label{related_work}
Feature Engineering can derive a new feature set from original features to improve downstream predictive performance~\cite{AFT1, AFT2, ying2024feature, gong2024neuro}.
Humans can manually reconstruct a feature set with domain knowledge and empirical experiences. These methods are explicit, traceable, and explainable, but are incomplete and time-consuming due to their heavy reliance on domain expertise and empirical insights.
Supervised transformations include exhaustive-expansion-reduction~\cite{expansion-reduction1, expansion-reduction2, expansion-reduction3, expansion-reduction4, expansion-reduction5}, iterative-feedback-improvement~\cite{envolution1, envolution2, envolution3, envolution4, envolution5, envolution6}, and AutoAI-based approaches~\cite{autoai1, autoai2, automl1, automl2, zhang2021automated, wever2021automl, bahri2022automl, wang2021autods}. 
Exhaustive-expansion-reduction methods use explicit~\cite{explicit} or greedy~\cite{greedy} techniques to construct feature cross, assessing their impact on downstream ML tasks and then proceeding with feature selection to keep crucial features. However, these approaches heavily rely on domain knowledge and labels, is hard to construct complex feature crosses, resulting in suboptimal performance.
Iterative-feedback-improvement methods leverage reinforcement learning, genetic algorithms, or evolution algorithms to formulate feature transformation as a discrete optimization task, searching for the optimal feature set iteratively. However, they face exponentially growing combination possibilities, are computationally costly, hard to converge, and unstable. 
AutoAI-based methods utilize AutoML~\cite{automl3, automl4} techniques to systematically explore various feature crosses in a structured, non-exhaustive manner. However, they are constrained by their inability to generate high-order features and their sensitivity to performance instability.
In contrast, unsupervised transformations like Principal Components Analysis (PCA)~\cite{PCA} don't rely on labels. However, PCA is based on a strong assumption of straight linear feature correlation, only uses addition/subtraction to generate new features, and leads to only dimensionality reduction rather than an increase.
To fulfill these gaps, we develop a measurement-pretrain-finetune paradigm, formulating Feature Transformation as an unsupervised continuous optimization task. This method constructs a continuous space with unsupervised graph contrastive learning, facilitating the generation of the best feature set. By decoupling FT from downstream tasks and avoiding ineffective discrete searches, our approach provides a more efficient solution.

\vspace{-0.1cm}
\section{Conclusion}
In this paper, we integrate graph, contrastive, and generative learning to develop a label-free generative framework for unsupervised feature transformation learning problems. 
There are three important contributions:
1) We develop an unsupervised measurement to evaluate feature set utilities, decoupling the framework from downstream tasks, and avoiding the evaluation of feature sets from lengthy and expensive experiments.
2) We regard a feature set as a feature-feature similarity graph and introduce an unsupervised training paradigm, embedding feature transformation knowledge into a continuous space through graph contrastive pre-training. 
3) We see feature transformation as a sequential generation task and develop an encoder-decoder-evaluator structure, which facilitates the identification of the optimal transformed feature set through gradient optimization. This structure enables the generation of optimal results, avoiding the need to explore exponentially growing possibilities of feature combinations in discrete space.
Extensive experiments validate the effectiveness, efficiency, traceability, and clarity of our framework. The superior experimental performance highlights its potential as a valuable approach for unsupervised generative feature transformation in large quantities of domains, such as bioinformatics, health care, finance analysis, etc.
\section{Acknowledgement}
This research was partially supported by the National Science Foundation (NSF) via the grant number: GR45865.
\bibliographystyle{ACM-Reference-Format}
\bibliography{ref}
\section{Appendix}

\subsection{RL data collector}
\label{appendix-RL}
Our perspective is to view reinforcement intelligence as a training data collector in order to achieve volume (self-learning enabled automation), diversity(exploration), and quality (exploitation).
To implement this idea, we design reinforcement agents to automatically decide how to perform feature crosses and generate new feature spaces. The reinforcement exploration experiences and corresponding accuracy will be used as training data. Specifically, the approach includes: 
1) \noindent\textbf{Multi-Agents: } We design three agents to perform feature cross: a head feature agent, an operation agent, and a tail feature agent. 
2)  \noindent\textbf{Actions: }  In each reinforcement iteration, the three agents collaborate to select a head feature, an operator (e.g., +,-,*,/), and a tail feature to generate a new feature. The newly generated feature is later added to the feature set for the next feature generation. 
3)  \noindent\textbf{Environment: } The environment is the feature space, representing an updated feature set. When three feature agents take actions to cross two features to generate a new feature and add the new feature to the previous feature set, the state of feature space (environment) changes. The state represents the statistical characteristics of the selected feature subspace. 
4)  \noindent\textbf{Reward Function: } The reward is an unsupervised feature set utility measurement proposed in methodology.
5)  \noindent\textbf{Training and Optimization: }  Our reinforcement data collector includes many feature cross steps. Each step includes two stages: control and training. In the control stage, each feature agent takes actions based on their policy networks, which take the current state as input and output recommended actions and the next state. The three feature agents will change the size and contents of a new feature space. We regard the new feature space as an environment. Meanwhile, the actions taken by feature agents generate an overall reward.  This reward will then be assigned to each participating agent. In the training stage, the agents train their policies via experience replay independently. 
The agent uses its corresponding mini-batch samples to train its Deep Q-Network (DQN), in order to obtain the maximum long-term reward based on the Bellman Equation.

\subsection{Pseudo-code of the algorithm}
\label{pseudo-code}
\begin{algorithm}
    \SetAlgoNoLine
    \LinesNumbered
    \SetKwInOut{Input}{Input}
    \SetKwInOut{Output}{Output}
    \Input{The original dataset $D = (X, y)$}
    \Output{The Optimal Transformed Feature set $\mathcal{F}^{*}$ }.
    Initialize the encoder $\phi$, decoder $\psi$ and evaluator $\theta$. \\
    \textbf{Unsupervised Measurement (Section 3.2):} \\
    Develop an unsupervised measurement (MDCG) to assess the utility of the feature set. \\
    Using RL and MDCG to collect $N$ training data $\Tilde{D}=\{\mathcal{F}_i, \mathcal{T}_i, s_i\}_{i=1}^N$

    \textbf{Graph Contrastive Pre-training (Section 3.3):} \\
    \For{$in ~pretrain-epoch$}
    {
        Construct feature-feature similarity graphs $\mathcal{F}_i \rightarrow \mathcal{G}_i$. \\
        Graph augmentation $\mathcal{G}_i \rightarrow (\mathcal{G}_{i,1}, \mathcal{G}_{i,2})$. \\
        Encode: $\mathbf{z}_{i,1} = \phi(\mathcal{G}_{i,1}), \mathbf{z}_{i,2} = \phi(\mathcal{G}_{i,2})$ \\
        Contrastive learning: $\mathcal{L}_{cl} = -\frac{1}{N}\sum_{n=1}^Nlog\frac{exp(sim(\mathbf{z}_{i, 1}, \mathbf{z}_{i, 2})/\tau))}{\sum_{i^*=1, i^* \neq i}^Nexp(sim(\mathbf{z}_{i, 1}, \mathbf{z}_{i^*, 2})/\tau)},$
    }

    \textbf{Multi-Objective Fine-tuning (Section 3.4):} \\
    \For{$in ~finetune-epoch$}
    {
        Encode: $\mathbf{\mathbf{z}} = \phi(\mathcal{G})$. \\
        Estimate: $\mathbf{m}, \mathbf{\sigma}$. \\
        Decode loss: $\mathcal{L}_{rec} = -logP_{\psi}(\mathcal{T}|\mathbf{z})$. \\
        Evaluate loss: $\mathcal{L}_{evt} = MSE(s, \theta(\mathbf{z}))$. \\ 
        Backward: $\mathcal{L} = \alpha\mathcal{L}_{evt}+\beta\mathcal{L}_{rec}$

    }
    Select top-$k$ transformed feature sets (graphs) $(\mathcal{G})^k$ from $\Tilde{D}$. \\
    Encode: $(\mathbf{z})^k= \phi((\mathcal{G})^k)$. \\
    Update $(\mathbf{z})^k$ with $\eta$ steps: $(\mathbf{z^+})^k = (\mathbf{z})^k + \eta*\frac{\partial\vartheta}{\partial{(\mathbf{z})^k}}$. \\
    Decode: $(\mathcal{T}^+)^k= \psi((\mathbf{z^+})^k)$. \\
    Optimal Transformed Feature Set: $\mathcal{F}^* = \arg\max\mathcal{S}((\mathcal{T}^+)^k[X])$.\\
    \caption{Entire Procedure}
    \label{pseudo-code-algo}
\end{algorithm}
The entire procedure algorithm is described as Algorithm~\ref{pseudo-code-algo}.

\subsection{Time complexity check with baselines}
\label{time-complexity-baselines}
\setlength{\tabcolsep}{2.5mm}{
\begin{table*}[tb]
\centering
\small
\caption{Time cost comparisons with baselines}
\vspace{-0.2cm}
\begin{tabular}{@{}cccccccccccc@{}}
\toprule\toprule
Dataset     & RDG(s) & PCA(s) & LDA(s) & ERG(s) & AFAT(s) & NFS(s) & TTG(s) & GRFG(s) & DIFFER(min) & MOTA(min) & \model(min) \\ \midrule
spectf      & 2.9    & 0.3    & 0.3    & 4.9    & 3.8     & 8.2    & 9.1    & 282     & 70.8        & 156.0     & 52.3      \\
svmguide3   & 7.0    & 0.2    & 0.8    & 10.3   & 8.8     & 7.1    & 8.3    & 373     & 68.1        & 191.8     & 59.7      \\
german      & 3.6    & 0.2    & 0.7    & 6.0    & 3.5     & 6.1    & 7.9    & 182     & 85.6        & 105.6     & 78.1      \\
openml\_616 & 27.6   & 0.2    & 0.2    & 34.1   & 2.6     & 18.1   & 24.7   & 936     & 75.6        & 430.1     & 62.1      \\
openml\_618 & 67.3   & 0.2    & 0.2    & 82.4   & 3.0     & 30.9   & 38.7   & 821     & 90.3        & 373.8     & 81.7      \\
openml\_620 & 33.4   & 0.2    & 0.2    & 42.1   & 2.1     & 17.0   & 22.6   & 1131    & 74.5        & 500.9     & 63.8      \\ \bottomrule\bottomrule
\end{tabular}
\vspace{-0.3cm}
\label{appendix-time}
\end{table*}}
We compared our method with baseline methods on 6 selected datasets over classification and regression tasks to report their time costs, as shown in Table~\ref{appendix-time}. The unsupervised baselines cost less time, but are less accurate. Compared with the supervised baselines, our method costs less time than DIFFER and MOTA and costs a bit more time than the other four supervised. However,
1). Feature transformation is not a data preparation step and not timing critical for most tasks;
2) The time cost increase (mins level) of training our method is acceptable. Compared with the time costs (days or months level) of human manual feature engineering, our method saves time;
3) Although our method costs a bit more time, it achieves much better feature engineering performance;
4) The inference time is small and it generates transformed features quickly in 0.6~1.7 s;
5) Based on such unsupervised and encode-decode generative design, we can use the strategy of pretraining a foundation model and then finetuning to reduce training time costs.

\subsection{The impact of different RL}
\setlength{\tabcolsep}{0.8mm}{
\begin{table}[tb]
\centering
\small
\caption{The impact of different RL.}
\vspace{-0.2cm}
\begin{tabular}{@{}cccc@{}}
\toprule\toprule
Dataset     & value iteration RL & Policy gradient RL & randomly generation \\ \midrule
spectf      & 0.856              & 0.856              & 0.823               \\
svmguide3   & 0.832              & 0.830              & 0.821               \\
german      & 0.723              & 0.726              & 0.715               \\
openml\_616 & 0.536              & 0.540              & 0.493               \\
openml\_618 & 0.616              & 0.612              & 0.520               \\
openml\_620 & 0.584              & 0.585              & 0.564               \\ \bottomrule\bottomrule
\end{tabular}
\vspace{-0.3cm}
\label{impact-rl}
\end{table}}
We conducted a thorough examination of the impact of training data generated by policy gradient and value learning reinforcement learning (RL) algorithms, as well as data generated randomly, as shown in Table~\ref{impact-rl}. Our findings indicate that training data produced using RL algorithms significantly outperforms randomly generated data in terms of enhancing model performance. However, it is important to note that despite the overall superiority of RL-generated data, the variations in performance between different RL designs are relatively limited. This suggests that while the choice of using RL over random generation is crucial for improving results, the specific type of RL algorithm employed may not drastically alter the final outcomes.

\subsection{Time complexity check of the different components in \model}
\label{app-diff}
\setlength{\tabcolsep}{4mm}{
\begin{table}[tb]
\centering
\small
\caption{Time complexity check of the different components in \model.}
\vspace{-0.2cm}
\begin{tabular}{@{}cccc@{}}
\toprule\toprule
Dataset     & RL(min) & Model training(min) & Inference(s) \\ \midrule
spectf      & 14.2    & 38.1                & 1.1          \\
svmguide3   & 23.6    & 36.1                & 1.1          \\
german      & 48.5    & 29.6                & 0.6          \\
openml\_616 & 22.5    & 39.8                & 1.3          \\
openml\_618 & 50.2    & 31.4                & 0.8          \\
openml\_620 & 34.3    & 29.5                & 0.7          \\ \bottomrule\bottomrule
\end{tabular}
\vspace{-0.3cm}
\label{impact-components}
\end{table}}

We reported the computational overheads of 1) RL data collection; 2) model training; 3) inference, as shown in Table~\ref{impact-components}. We found that the additional time cost incurred by {\model} occurs during the data collection and model training phases. However, once the model converges, {\model}'s inference time is significantly reduced. The potential driving factors are that the RL-based collector spends more time gathering high-quality data, and the sequence formulation across the entire feature space increases the learning time cost for the sequence model. Nonetheless, during the inference phase, {\model} outputs the entire feature transformation at once, resulting in a very short inference time.

\end{document}